# Effective and efficient structure learning with pruning and model averaging strategies

Anthony C. Constantinou, Yang Liu, Neville K. Kitson, Kiattikun Chobtham, and Zhigao Guo.

Bayesian Artificial Intelligence research lab, School of Electronic Engineering and Computer Science, Queen Mary University of London (QMUL), London, UK, E1 4NS.
E-mails: a.constantinou@qmul.ac.uk, yangliu@qmul.ac.uk, n.k.kitson@qmul.ac.uk, k.chobtham@qmul.ac.uk, and zhigao.guo@qmul.ac.uk.

**Abstract:** Learning the structure of a Bayesian Network (BN) with score-based solutions involves exploring the search space of possible graphs and moving towards the graph that maximises a given objective function. Some algorithms offer exact solutions that guarantee to return the graph with the highest objective score, while others offer approximate solutions in exchange for reduced computational complexity. This paper describes an approximate BN structure learning algorithm, which we call Model Averaging Hill-Climbing (MAHC), that combines two novel strategies with hill-climbing search. The algorithm starts by pruning the search space of graphs, where the pruning strategy can be viewed as an aggressive version of the pruning strategies that are typically applied to combinatorial optimisation structure learning problems. It then performs model averaging in the hill-climbing search process and moves to the neighbouring graph that maximises the objective function, on average, for that neighbouring graph and over all its valid neighbouring graphs. Comparisons with other algorithms spanning different classes of learning suggest that the combination of aggressive pruning with model averaging is both effective and efficient, particularly in the presence of data noise.

**Keywords:** *Bayesian networks, directed acyclic graphs, Markov equivalence, noisy data, probabilistic graphical models*.

## 1. Introduction

A Bayesian Network (BN) is a probabilistic graphical model consisting of nodes and arcs, represented by a Directed Acyclic Graph (DAG) and its corresponding conditional distributions. Learning a BN model involves determining its graphical structure and parameterising its conditional distributions. Discovering an accurate structure efficiently represents a machine learning (ML) problem that is generally NP-hard, and solutions to this problem include algorithms spanning different classes of learning. The two main learning classes are constraint-based and score-based learning, where the former primarily relies on statistical tests of conditional independence and the latter involves exploring and scoring the search space of graphs. Hybrid algorithms also exist that combine these two classes of learning. Kitson et al (2021) provide a review of the many different algorithms that fall within different classes of learning.

Algorithms that rely on constraint-based learning are statistical in nature. They typically start from a fully connected undirected graph and perform conditional independence tests to remove edges, and to orientate as many of the edges as possible. The PC algorithm (Spirtes et al., 2000) is probably the most widely known constraint-based algorithm. This is because it represents a set of interesting learning strategies that have been adopted by many subsequent constraint-based algorithms. Whilst controversial, constraint-based learning is often stated to discover graphs in which directed edges represent causal relationships, and undirected edges represent causation in which the direction of causation cannot be determined solely by observational data. These claims, however, assume a set of conditions about the input data that only clean synthetic data are likely to satisfy (Spirtes et al., 2000).

In contrast, the class of score-based learning combines search methods from ML with objective functions. The search methods determine how the algorithm traverses the search space of graphs, whereas the objective function assigns an evaluation score to each graph visited. The highest scoring graph discovered is then returned as the preferred graph. Because the AI field covers several search algorithms, and because there are different objective functions, there is a much greater variety in score-based algorithms compared to constraint-based algorithms. In the context of BN structure learning, a major distinction between score-based algorithms involves exact and approximate learning, where the former guarantees to return the graph that has the highest objective score in the search space (although the search space is often constrained by a maximum in-degree) while the latter does not provide this guarantee.

Exact learning solutions tend to be computationally expensive, and this restricts their application to low dimensional problems that involve lower number of parameters. On the other hand, approximate solutions naturally scale to higher dimensional problems that may contain hundreds and occasionally thousands of variables. Whilst returning the graph that maximises a given objective function represents an interesting ML exercise, it is important to clarify that the maximal objective graph is often not the ground truth causal graph (for details refer to subsection 3.3). This means there might be cases where a high scoring graph represents a more accurate causal graph than the highest scoring graph, and this naturally diminishes the incentive for pursuing the highest scoring graph[1] at the expense of efficiency. This paper builds on this **hypothesis** by investigating an efficient and effective model averaging solution.

***Heuristic search*** typically involves approximate learning solutions that employ variants of greedy search, such as hill-climbing and tabu, to explore the search space of graphs. In the context of structure learning, a hill-climbing search explores neighbouring graphs by performing graph modifications such as edge additions, edge removals, and edge reversals, and moves to the neighbouring graph that increases the objective score the most. Tabu search is a popular variant of hill-climbing that includes visits to neighbouring graphs that minimally decrease the score in an effort to escape local maxima solutions on which hill-climbing is known to get stuck on.

The earliest Hill-Climbing (HC) and TABU structure learning algorithms are likely to be those described by Bouckaert (1994; 1995) and Heckerman et al. (1995), even though

[1] The set of 'high scoring' graphs can be enormous. Because of this, it is not possible to know whether the preferred high scoring graph will be sufficiently accurate, or more accurate compared to the highest scoring graph, in terms of distance from the ground truth graph.

Cooper and Herskovits used a form of greedy search for their score-based K2 algorithm a few years earlier (1992). Since then, several subsequent score-based and hybrid structure learning algorithms rely on these early versions of greedy heuristic search. In addition to the score-based K2 algorithm, some other well-established algorithms include the score-based Greedy Equivalence Search (GES) algorithm by Chickering (2002) which greedily searches the space of Markov equivalence DAGs, also known as *essential* graphs or Complete Partial DAGs (CPDAGs), and the hybrid learning Max-Min Hill-Climbing (MMHC) algorithm by Tsamardinos et al. (2006) which constructs a skeleton using constraint-based learning and then performs Tabu search where arc additions are constrained to the edges present in the skeleton graph.

***Pruning*** the search space of graphs represents a particularly important strategy for improving the efficiency of exact learning algorithms in particular. This is because, in the absence of pruning, an exact algorithm would need to perform exhaustive search to guarantee the discovery of the optimal graph. Exhaustive search represents an unrealistic approach since the number of possible graphs grows super-exponentially in the number of variables. In other words, applying exhaustive search to the unpruned search space of graphs can be prohibitive even for networks that contain less than 10 variables (Kitson et al., 2021). This makes pruning a requirement for exact learning. Note that the pruned edge-set corresponds to the edges preserved following pruning, whereas the edges pruned off correspond to the edges that will not be visited due to pruning.

An important distinction between pruning strategies involves whether the pruning is *sound* or not. That is, sound pruning ensures the pruned search space of graphs contains the optimal graph. Research into sound pruning has been important in the development of exact search. Well-established exact learning solutions include integer programming approaches such as GOBNILP by Cussens (2011), and combinatorial optimisation approaches such as Branch-and-Bound by de Campos (2009). Both these approaches employ effective versions of sound pruning and allow exact learning to scale to tens of variables, which is still insufficient since their application is restricted to relatively low dimensionality problems. Importantly, the efficiency of these solutions is dependent on the assumptions made about the underlying distributions that represent the data. For example, if we assume that the data are continuous and normally distributed, the algorithm will converge much faster compared to, for example, assuming discrete data, but likely at the expense of accuracy if not all data are distributed as assumed.

Because pruning strategies require that a set of local scores is pre-computed before a pruning decision is determined as well as generally before an algorithm enters the actual structure learning phase, this process might introduce significant upfront computational complexity. This additional computational complexity is much more evident for approximate learning algorithms that explore a minor part of the search space. However, because pruning for approximate learning algorithms need not to be sound, it can be more aggressive such that it minimises both the optimisation/pre-processing complexity as well as the structure learning complexity considerably. For example, Guo and Constantinou (2020) show that pruning Candidate Parent Sets (CPSs) by removing those with relatively low local scores leads to marginal reductions in structure learning accuracy in exchange for important reduction in the computational complexity, thereby easing the application of structure learning to data sets that contain thousands of variables. Other pruning strategies can be specific to a particular objective function. For example, the ASOBS algorithm by Scanagatta et. al. (2015) uses an approximate BIC score, which they call BIC*, to perform a more aggressive form of pruning where pruning conditions can be used to prune supersets of parents without having to score those supersets. These pruning conditions tend to remove large portions of the search space, and this makes the algorithm considerably more efficient at a relatively small cost in accuracy.

***Model averaging*** aims to reduce inconsistencies in the learnt output. It may involve returning the average output over multiple outputs produced by different ML algorithms, or the average output over multiple candidate outputs as determined by a single ML algorithm. In the context of structure learning, unlike model selection which involves returning the single best graph discovered, model averaging typically involves returning an output that represents some weighted average across a set of high-scoring graphs. Importantly, this means that model averaging produces an approximate graph that will almost never match the graph produced by exact learning. One of the earliest papers that discuss the difference between model selection and model averaging in this context is the work by Madigan et al. (1996), where they argue that model averaging should be applied to a set of CPDAGs rather than to a set of DAGs. Recent related works include those by Chen and Tian (2014) who implemented an algorithm to return the $k$-best equivalence classes of BN structure for model averaging, by Goudie and Mukherjee (2016) who describe a Gibbs sampler for learning DAGs that involves averaging across a set of DAGs that satisfy a set of conditions, and by Kuipers et al (2021) who propose a hybrid learning algorithm that samples DAGs from the posterior distribution to reduce the complexity of MCMC and enable full Bayesian model averaging for large networks.

The algorithm described in this paper combines the three features discussed above. Namely, existing and novel pruning strategies are first applied to the search space of graphs before the algorithm greedily explores the search space with model averaging. The paper is structured as follows: Section 2 describes the MAHC algorithm along with preliminary technical information, Section 3 presents and discusses the results, and we provide our concluding remarks and directions for future research in Section 4.

# 2. The MAHC algorithm (with preliminaries)

The Model Averaging Hill-Climbing (MAHC) algorithm can be viewed as a variant of the classic HC algorithm with two extensions. In brief, the first extension involves pre-processing some of the local objective scores and applying pruning to the search space of DAGs. The outcome of the pre-processing step will be a set of arcs pruned off, in addition to a set of local scores pre-processed that can be reused during the structure learning phase. The second extension involves applying model averaging over the hill-climbing search space. The subsections that follow cover these two extensions in turn.

## *2.1. Pruning the search space of graphs*

As discussed in the Introduction, pruning in structure learning is both more necessary, and hence more common, in exact algorithms that rely on combinatorial optimisation rather than approximate heuristics. We first explain how traditional pruning is applied to these combinatorial optimisation problems, and then describe how we have modified traditional pruning to become suitable for heuristic search.

We denote the set of discrete variables by uppercase letter $V$, the CPS $j$ of variable $V_i$ by $CPS_{i,j}$ where $i$ iterates over all variables and $j$ iterates over the CPSs of $V_i$, and $S_{i,j}$ corresponds to the objective score of $CPS_{i,j}$. We use the standard Bayesian Information Criterion (BIC) as the objective score in the traditional form of:

$$\text{BIC} = LL(G|D) - \left(\frac{\log_2 N}{2}\right) p \tag{1}$$

where $LL$ is the Log-Likelihood score of graph $G$ given input data set $D$, $N$ is the sample size of $D$, and $p$ is the number of free parameters in $G$ obtained by:

$$p = \sum_{i}^{|V|} \left( (|V_i| - 1) \prod_{k}^{|CPS_i|} |V_k| \right) \tag{2}$$

where $|V|$ is the size of set $V$, $|V_i|$ is the number of states in $V_i$, $|CPS_i|$ is the number of parents of $V_i$, and $|V_k|$ is the number of states of each parent $V_k$ that is a member of $CPS_i$.

When structure learning operates over CPSs, the simplest pruning rule states that, assuming $CPS_{1,1}$ has local score $S_{1,1}$ and the $CPS_{1,2}$ has local score $S_{1,2}$, if $CPS_{1,2}$ is a subset of $CPS_{1,1}$ and $S_{1,2} \geq S_{1,1}$ then we can safely prune off $CPS_{1,1}$ without affecting the final output of the graph. This is because if the superset $CPS_{1,1}$ is valid and does not lead to cycles then so will its subset $CPS_{1,2}$, and since $CPS_{1,1}$ offers no better score than its subset $CPS_{1,2}$ then $CPS_{1,1}$ will never form part of the optimal graph.

Importantly, the above pruning rule is sound in the sense that it guarantees not to prune off CPSs that could have been included in the optimal graph. Sound pruning rules are particularly important when applied to exact algorithms. Because the algorithm we investigate in this paper is approximate, the incentive for pruning to be sound is relaxed. In other words, we can consider more aggressive forms of pruning that significantly reduce computational complexity in exchange for a small decrease in the expected objective score of the learnt graph. In fact, stronger forms of pruning are generally desired when performing model averaging, since the number of models explored with model averaging can be orders of magnitude higher than performing model selection.

Because the proposed algorithm explores the search space of graphs via heuristic search, it makes sense for the pruning to be applied on edges rather than on CPSs. Therefore, we have implemented the traditional optimisation-based pruning rule to greedy search, with the first difference being that, when a pruning condition is met, the pruning is applied to the edge rather than just to the CPS. For example, in assessing the CPSs of node $A$, denoted $CPS_{A,j}$, if $CPS_{A,1} = \{B\}$ (i.e., with parent $B$) is found to have a higher local score than $CPS_{A,2} = \{B, C\}$ (i.e., with parents $B$ and $C$), then in this study we remove all CPSs containing $C$, which means that each pruning decision may remove multiple CPSs rather than just one, such as $CPS_{A,2} =$ $\{B, C\}$ in this example. This cancels the guarantee for sound pruning in exchange for improved computational efficiency. This leads to the definition of **Pruning rule**.

> **Pruning rule:** Assuming $CPS_{1,1}$ and $CPS_{1,2}$ have corresponding scores $S_{1,1}$ and $S_{1,2}$, if $CPS_{1,1} \subset CPS_{1,2}$ and $S_{1,1} \geq S_{1,2}$, then the parents resulting from set subtraction $CPS_{1,2} - CPS_{1,1}$ are pruned off. Note that any edges pruned off apply to CPSs of all sizes.

As mentioned in the Introduction, an important limitation of pruning strategies is that, before they make pruning decisions, they typically require a set of local scores to be pre-processed, up to a given maximum in-degree. Because part of those pre-computed scores will never be visited during structure learning, this impacts the efficiency of the algorithm as well as memory allocation, depending on the pruning strategy and the structure learning algorithm. In the case of exact learning, this limitation is not as pronounced since all local scores would need to be computed in the absence of pruning anyway, and absence of pruning implies exhaustive search over all possible CPSs, bounded by the maximum in-degree. In this paper, however, we apply pruning to hill-climbing search. Because hill-climbing explores a tiny fraction of the search space, it is reasonable to expect that a pruning strategy would pre-process a large portion of local scores that would otherwise not have been computed.

To minimise this potentially significant added computational complexity, we apply **Pruning rule** dynamically while pre-processing the local scores that are used to determine the edges pruned off, and this is similar to the pruning strategy proposed by Scanagatta et. al. (2015). Specifically, for a given child node, pre-processing of supersets will not consider CPSs containing parents that were pruned off during pre-processing of their subsets. In other words, applying **Pruning rule** dynamically while pre-processing CPSs implies that we are not going to pre-process scores which are affected by "edge pruning", since we are not going to need them during structure learning.

Because the number of possible CPSs is generally enormous, applying **Pruning rule** to all the possible CPS combinations makes it highly likely that some true edges will be pruned off. To deal with this issue, we introduce constraints that limit the application of pruning to a certain set of CPSs. That is, assuming the maximum in-degree is 3 for the pre-processing (and not necessarily for the structure learning) step that determines the edges pruned off, we want to constrain **Pruning rule** such that it will visit a) all possible empty CPSs (one for each node), b) all possible single-parent CPSs, c) the valid dual-parent CPSs that include the highest scoring parent, and d) the valid triple-parent CPSs that include the two highest scoring parents. More specifically, consider $B$ and $C$ are respectively the highest and second highest scoring parents of $A$ (and higher than the empty CPS), identified after the algorithm has pre-processed all single-parent CPSs. The pre-processing of CPSs with set sizes 2 and 3 will then involve iterations over $V_i$ constrained over $CPS_{A,j} = \{B, V_i\}$ where $V_i \neq A, B$, and constrained over $CPS_{A,j} = \{B, C, V_i\}$ where $V_i \neq A, B, C$, respectively. This means that not all CPSs of size 2 and 3 will be visited during pre-processing. In other words, while **Pruning rule** is applied to all empty and single-parent CPSs (i.e., CPSs of size 0 and 1), its application on CPSs of size 2 and 3 is restricted to a specific set of CPSs. This restriction on CPSs of size 2 and 3 is generalised as **Pruning rule (with constraints)**.

> **Pruning rule (with constraints):** Each CPS corresponds to a node $i$ that is part of $V$, denoted as $V_i$, and each $V_i$ has $|V|-1$ possible parents ranked by $l^{\text{th}}$ highest score. Consider that the first and second highest valid[2] scoring parents of $V_i$ are $CPS_{i,l=1}$ and $CPS_{i,l=2}$ respectively; e.g., $CPS_{i,l=1}$ has the highest score as a CPS of size one (single parents) of $V_i$. When **Pruning rule** is executed on CPSs of size two for node $V_i$, it is only applied to CPSs that contain $\{CPS_{i,l=1}, V_k\}$ and iterate over $k$, where $V_k \notin \{CPS_{i,l=1}, V_i\}$. Similarly, when executed on CPS of size three, it will be restricted to CPSs that contain $\{CPS_{i,l=1}, CPS_{i,l=2}, V_k\}$ iterating over $k$, where $V_k \notin \{CPS_{i,l=1}, CPS_{i,l=2}, V_i\}$. In other words, for CPS sizes greater than 1, pruning is only applied to the CPSs that include the $n-1$ highest scoring valid parents.

In this paper, the entire pre-processing step is restricted to a maximum in-degree of 3, as in GOBNILP (refer to Table A1). However, it is important to clarify that the arcs pruned off are dependent on, and local scores pre-processed will be up to, a given maximum in-degree that does not need to be equal to the maximum in-degree assumed during structure learning. This implies that two hyperparameters are needed to specify the maximum in-degree; one for the pre-processing step and another for the structure learning step. These details are discussed in subsection 2.2.

**Example:** Consider the classic Asia network depicted in Fig 1. The network consists of eight nodes and has a maximum in-degree of 2. The number of CPSs to be pre-processed, in the absence of dynamic pruning, can be measured as follows:

$$\sum_{m=0}^{MID} \frac{|V|!}{m!\,(|V|-m)!}(|V|-m) \tag{3}$$

where $|V|$ is the number of variables as defined in Equation 2, $MID$ is a given maximum in-degree, and $m$ represents the size of CPSs at each in-degree iteration. Assuming $MID = 3$ for pre-processing, then:

- At iteration $m = 0$ the number of CPSs generated that contain zero parents is 8. At this iteration, the number of CPSs generated is also equivalent to $|V|$.

- At iteration $m = 1$ the number of CPSs generated that contain one parent is 56. At this iteration, the number of CPSs is also equivalent to:

$$\frac{|V|(|V|-1)}{1!}.$$

[2] It is possible for one of the highest scoring parents to be pruned off during pre-processing. This can happen when pre-processing CPSs of at least size 2. When this happens, the next available highest scoring parent takes the place, in the ladder of highest scores for a given node, of the parent that has been pruned off.

- At iteration $m = 2$ the number of CPSs generated that contain two parents is 168. At this iteration, the number of CPSs is also equivalent to:

$$\frac{|V|(|V|-1)(|V|-2)}{2!}.$$

- At iteration $m = 3$ the number of CPSs generated that contain three parents is 280. At this iteration, the number of CPSs is also equivalent to:

$$\frac{|V|(|V|-1)(|V|-2)(|V|-3)}{3!}.$$

This brings the total number of CPSs that can be potentially pre-processed, bounded by maximum in-degree of 3, to 512. Note this would be the number of CPSs pre-processed *without* dynamic pruning applied to the pre-processing step. Whilst the number of CPSs in this example is tiny, it is important to reiterate that the number of CPSs scales up super-exponentially in the number of variables when maximum in-degree is ≥3, and that the application of exact algorithms (with pruning) is generally restricted to data sets that contain tens of variables.

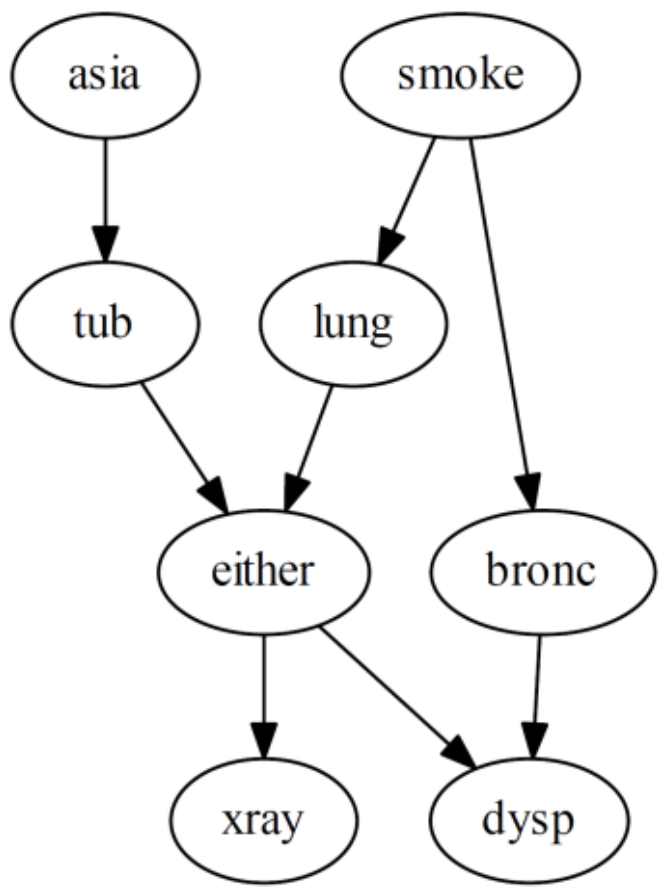

**Figure 1.** The true DAG of the classic Asia network.

Table 1 presents the local BIC scores generated for selected CPSs of the Asia network. The CPSs selected are those that capture all pruning decisions. The local BIC scores highlighted in red represent pruning decisions determined by **Pruning rule (with constraints)**, and the variables highlighted in red represent the parents of the corresponding child node that is pruned off. For example, reading along the row for *asia* in Table 1, we see that the highest scoring CPS for variable *asia* is the one with no parents. This means that no variable exists in the data that can serve as a parent of *asia* and increase the local BIC score relative to the score of the empty set (i.e., no parent). In the case of *asia*, all CPSs that contain two or three parents are pruned off before visited due to **Pruning rule**, since all edges entering *asia* are pruned off while pre-processing their subsets (i.e., empty and single parent CPSs). As shown in Fig 1, while this pruning decision happens to be correct in practice since *asia* has no parents in the true graph, we would expect *tub* to produce the highest score given the ground truth graph.

A different example involves the node *either*, which has two parents in the true graph. In this case, the comparison between CPSs that contain zero and one parent leads to the decision to prune off *asia* and *bronc* as parents of *either*. When the pre-processing moves to CPSs that contain two parents, further two variables are pruned off due to **Pruning rule (with constraints)**, and these are *smoke* and *dysp*. This is because when *smoke* and *dysp* are paired with the highest scoring parent, *lung*, they both return a lower score compared to the score returned by *lung* alone. Lastly, when pre-processing moves to CPSs that contain three parents, **Pruning rule (with constraints)** prunes off the *xray* variable. This is because the CPS $\{\mathrm{tub}, \mathrm{lung}, \mathrm{xray}\}$ produces a lower score than its subset $\{\mathrm{tub}, \mathrm{lung}\}$ (not shown in Table 1). Notice that, in this case, the *xray* variable that is pruned off was, in fact, the second highest scoring parent of *either*. Once more, these pruning decisions happened to be consistent with the true graph depicted in Fig 1.

In the above example which assumes a sample size of $10^3$, the pruning strategy has pre-processed eight empty CPSs, 56 single-parent CPSs, 48 valid dual-parent CPSs, and nine valid triple-parent CPSs. The total number of CPSs pre-processed is 121. Note that this is the number of CPSs pre-processed *with* pruning, and it is considerably smaller than the corresponding number of 512 CPSs (described above) that would have been pre-processed in the absence of pruning. Moreover, the CPSs pruned off will always be those which contain a higher number of parents and require more processing time, and this makes dynamic pruning even more effective. With reference to Table 1, the pre-processing of single-parent CPSs led to 28 pruning decisions, the pre-processing of dual-parent CPSs led to 12 pruning decisions, and the pre-processing of triple-parent CPSs led to five pruning decisions.

Recall that, in this paper, we consider each pruning decision to correspond to an arc pruned off, and subsequently will not be considered during structure learning. Therefore, a total of 45 arcs are forbidden out of the possible 56 arcs that could have been added to the Asia network which consists of eight nodes; i.e., there are $|V|(|V|-1)$ possible arcs in a network with $|V|$ nodes. This leaves just 11[3] arcs to be explored during the structure learning phase, which we describe in subsection 2.2. Algorithm 1 provides the pseudocode for the pre-processing phase with dynamic pruning. This also means that the theoretical complexity of the pruning phase, in terms of the number of CPSs visited, is

$$f(|V|) = O\left(\sum_{m=0}^{M_P} \frac{|V|!}{m!\,(|V|-m)!}(|V|-m) - PR_{M_P}\right) \quad (4)$$

where $|V|$ is the number of variables as defined in Equation 2, $M_P$ is a given maximum in-degree pruning hyperparameter, and $m$ represents the size of CPSs at each in-degree iteration as defined in Equation 3, and $PR_{M_P}$ represents the number of CPSs pruned off by **Pruning rule (with constraints)** given $M_P$. Subsection 3.5 includes empirical investigation of the time complexity of this pruning phase.

[3] These are either → tub, lung → smoke, bronc → smoke, either → lung, smoke → bronc, dysp → bronc, tub → either, lung → either, either → xray, bronc → dysp, and either → dysp.

**Algorithm 1:** Pre-processing of Candidate Parent Sets (CPSs) with dynamic pruning.

**Input:** data set $D$, empty score-set $S$ for each CPS $j$ of variable $i$ in data set $D$, maximum in-degree $M_P$.
**Output:** A set $F$ containing forbidden edges that have been pruned off, and a set $C$ containing the local BIC scores of each CPS pre-processed.

**for** up to in-degree $M_P$ **do**
  **for** each variable $i$ in $D$ **do**
    **for** each CPS $j$ **do**
      **if** CPS $j$ does not violate $F$ **then**
        **if** $|\mathrm{CPS}|$<2 **or if** $|\mathrm{CPS}|$>1 **and** CPS $j$ does not violate **Pruning rule (with constraints) then**
          compute objective score $S_{i,j}$ and add it to set $C$.
        **end if**
      **end if**
    **end for**
  **end for**
  Apply **Pruning rule** on $C$ and add any edges pruned off in set $F$.
**end for**
**end for**

**Table 1.** An example of the pruning rules applied to the Asia network (note that not all possible CPSs are shown). The CPSs shown here are those that capture all arcs pruned off. The results assume clean synthetic data with a sample size of $10^3$. The values in the table represent the local BIC score as determined by Equation 1. The local BIC scores in red with yellow backcolour represent pruning decisions and correspond to the variables in red which represent the parents pruned off due to those scores. Moreover, P indicates that the CPS was not visited due to an edge that was pruned off at a lower in-degree, whereas n/a indicates impossible local graph.

| Child node \ CPS | {} | {asia} | {tub} | {smoke} | {lung} | {bronc} | {either} | {xray} | {dysp} | {either, xray} | {bronc, smoke} | {bronc, dysp} |
|---|---|---|---|---|---|---|---|---|---|---|---|---|
| asia | -92.3 | n/a | -95.2 | -97.3 | -96.3 | -97.3 | -97.3 | -97.3 | -96.0 | P | P | P |
| tub | -79.1 | -81.9 | n/a | -83.4 | -83.8 | -81.2 | -48.3 | -54.9 | -80.8 | -58.1 | P | P |
| smoke | -1004.8 | -1009.8 | -1009.1 | n/a | -978.4 | -924.6 | -985.1 | -995.6 | -967.8 | P | n/a | -928.7 |
| lung | -332.4 | -336.4 | -337.2 | -306 | n/a | -335.7 | -45.5 | -130.8 | -300.6 | -55.3 | P | P |
| bronc | -998.6 | -1003.6 | -1000.7 | -918.4 | -1001.8 | n/a | -1000.5 | -1002.5 | -614.7 | P | n/a | n/a |
| either | -363.4 | -368.3 | -332.6 | -343.6 | -76.5 | -365.3 | n/a | -127.4 | -329.3 | n/a | P | P |
| xray | -504.9 | -509.9 | -480.8 | -495.6 | -303.3 | -508.8 | -268.9 | n/a | -494.2 | n/a | P | P |
| dysp | -993.5 | -997.2 | -995.3 | -956.5 | -961.7 | -609.6 | -959.4 | -982.8 | n/a | P | -613.7 | n/a |

| Child node \ CPS | {either, smoke} | {either, dysp} | {either, tub} | {either, lung} | {lung, smoke} | {lung, dysp} | {lung, bronc, either} | {lung, bronc, xray} | {tub, lung, xray} | {bronc, either, lung} | {bronc, either, xray} |
|---|---|---|---|---|---|---|---|---|---|---|---|
| asia | P | P | P | P | P | P | P | P | P | P | P |
| tub | P | P | P | P | P | P | P | P | n/a | P | P |
| smoke | P | P | P | P | P | P | -923.0 | -922.5 | P | P | P |
| lung | -48.6 | -55 | P | n/a | n/a | n/a | n/a | n/a | n/a | n/a | P |
| bronc | P | P | P | P | P | P | n/a | n/a | P | n/a | n/a |
| either | n/a | n/a | n/a | n/a | -86.2 | -83.7 | n/a | P | -39.9 | n/a | n/a |
| xray | -278.5 | -277.7 | -278.7 | -278.7 | P | P | P | n/a | n/a | P | n/a |
| dysp | P | n/a | P | P | P | n/a | P | P | P | -581.7 | -581.2 |

### *2.2. Determining the best neighbouring graph via model averaging*

Once the pre-processing phase is completed and the set of edges that can be considered for structure learning is determined, the algorithm moves to the structure learning phase. The search space of graphs is explored using a novel hill-climbing variant adjusted to model averaging. In the context of structure learning, traditional hill-climbing involves starting from an empty DAG and exploring neighbouring[4] DAGs via arc additions, deletions and reversals, and traverses to the neighbouring DAG that maximises the objective function. The search stops when no neighbouring DAG exists that further increases the objective score, and the graph found that maximises the objective score is returned as the preferred graph.

Unlike traditional hill-climbing where the best neighbouring graph is determined solely by its objective score, the variant we have implemented in this study determines the best neighbouring graph by its score and the scores of all its valid neighbouring graphs. A valid neighbour represents a DAG that is consistent with the pruning decisions executed during the pre-processing step; i.e., it does not contain any edges pruned off. This approach can be viewed as an extension of hill-climbing search where the depth level of neighbouring graphs visited per hill-climbing iteration is increased by one level. Each neighbouring graph is assigned an average objective score, determined over the score of the given neighbouring graph and the scores of its valid neighbouring graphs. Therefore, this extended version traverses to the neighbouring graph that maximises the average score over a set of scores, rather than maximising a single score.

Note that unlike traditional model averaging which involves averaging over a set of graphs, the model averaging approach employed in this study involves averaging a set of model selection scores (i.e., BIC scores), and each average score is assigned to a single graph being explored. The formal description of this modification is provided by **Modification 1** and **Modification 2**, for search and score respectively.

**Modification 1 (Search):** Given a candidate DAG $G$, traditional hill-climbing involves visiting each neighbouring graph $G_n$ of $G$ at each hill-climbing iteration. In the extended version, we modify search such that each hill-climbing iteration involves not only visiting each neighbouring graph $G_n$ of $G$, but also each neighbouring graph $G_{nn}$ of $G_n$ (i.e., $G_{nn}$ is the neighbouring graph of the neighbouring graph of $G$).

**Modification 2 (Score):** Given a candidate DAG $G$, traditional hill-climbing moves to the neighbouring graph that maximises $S(G_n)$ given a set of scores $S_n$ that consists of multiple $S(G_n)$. In other words, traditional hill-climbing searches for the maximum objective score $S(G_n)$ across neighbouring scores. In the extended version, hill-climbing moves to the neighbouring graph that returns $\max\left(\overline{S(G_n, G_{nn})}\right)$ given a set of scores $S_n$ that consists of multiple $\overline{S(G_n, G_{nn})}$. In other words, the extended version searches for the highest average objective score $\overline{S(G_n, G_{nn})}$, each of which corresponds to a neighbouring graph $G_n$ and the scores of all its valid neighbouring graphs $G_{nn}$.

[4] Note that an arc added at an iteration may be unconnected to the DAG from the previous iteration – that is, the DAG may 'grow' from several distinct initial edges.

Note that if we were to retrieve $\max\left(\max\left(S(G_n, G_{nn})\right)\right)$ instead of $\max\left(\overline{S(G_n, G_{nn})}\right)$; i.e., retrieving the highest score across the two hill-climbing walks rather than the highest average, this would likely lead to higher objective scores and likely higher graph-based scores in the presence of clean data. Because this paper also considers the case of noisy data, which represents a more realistic scenario in practice, this diminishes the incentive to maximise score fitting. This is because the highest scoring graph will be the one that best fits the data *and* the noise. On this basis, we explore how an approximate model averaging approach, that averages over a set[5] of scores, compares to traditional approximate and exact learning strategies that traverse towards the highest score.

Fig 2 illustrates an example of the extended hill-climbing heuristic we employ in this paper. The figure is based on the Asia network previously covered in subsection 2.1, as well as on the scores depicted in Table 1. The illustration starts from the penultimate candidate graph depicted on the top left corner, which represents the second-to-last highest scoring graph as identified by MAHC. From that candidate graph, the figure focuses on illustrating the step that identifies the highest scoring neighbouring graph which represents the final output graph. Specifically, the neighbouring graph that reverses $bronc \rightarrow smoke$ into $smoke \rightarrow bronc$ represents the final output graph, and whose objective score is determined by its own BIC score and the BIC scores of its 11 valid neighbouring graphs. A valid neighbouring graph is one that satisfies acyclicity as well as all the pruning conditions (the pruning conditions of this example can be found in Table 1). Each arc reversal is highlighted in green and each arc removal in red.

In Fig 2, none of those 11 neighbouring graphs involves adding a new arc, and this is because, in this example, it happened to be the case that all other possible arc additions are pruned off and could not have been considered in determining the average score (refer to Table 1 and to the footnote in subsection 2.1 for details). The final output graph has an average BIC score -3357.73, and this score represents the average BIC score of the 12 graphs; i.e., the neighbouring graph of the penultimate graph and its 11 neighbouring graphs. The scores of the final output graph can be found in Table B1. However, note that the BIC scores depicted in Table B1 reflect the score of the learnt graph and not the model averaging BIC score illustrated here.

The pseudocode of MAHC is provided by Algorithm 2. As mentioned in subsection 2.1, the maximum in-degree can vary between pre-processing for pruning and structure learning. MAHC takes two hyperparameter inputs to reflect this, as illustrated by Algorithm 2, where $M_P$ determines the maximum in-degree regarding the size of CPSs visited during pre-processing as described in subsection 2.1, and $M_S$ determines the maximum in-degree for structure learning as described above. This makes the algorithm more flexible and enables more reasonable restrictions of max in-degree to be applied to each phase. This is because it would be unreasonable to pre-process CPSs up to a high maximum in-degree, since this will substantially increase the number of CPSs pre-processed and which are only occasionally visited during structure learning. On the other hand, allowing heuristic search to explore a higher number of parents than those pre-processed had relatively small impact on time complexity. This is because hill-climbing tends to visit few CPSs with in-degree greater than 3, which end up having limited impact on computational complexity compared to pre-computing all valid CPSs. As indicated in Table A1, the default hyperparameter inputs for maximum in-degree are '3' for pre-processing and '8[6]' for structure learning. In other words, the pruning phase computes and saves all valid CPS scores up to in-degree 3, whereas the structure learning phase is allowed to explore denser graphs up to in-degree 8. Therefore, it would be reasonable for $M_S \geq M_P$ to always hold, to ensure that all the scores for the CPSs pre-processed during the pruning phase are reused during the structure learning phase.

[5] The number of neighbours is the same for every immediate neighbour at a given iteration. However, because MAHC averages graphs over two hill-climbing walks, it is possible that score averages are derived from neighbouring sets of graphs of different size. This is because two hill-climbing walks (i.e., when visiting the neighbour of the neighbour) can involve different numbers of cyclic graphs, which are excluded from the averaging process. Because the difference in the number of valid graphs visited by MAHC is trivial for a given best graph, we compute the average scores without worrying about possible differences in the number of neighbours visited to produce those averages.

[6] Note that this is the hyperparameter default at implementation level. In terms of the results presented in this paper, setting MID to 8 returns the same results as leaving MID unbounded.

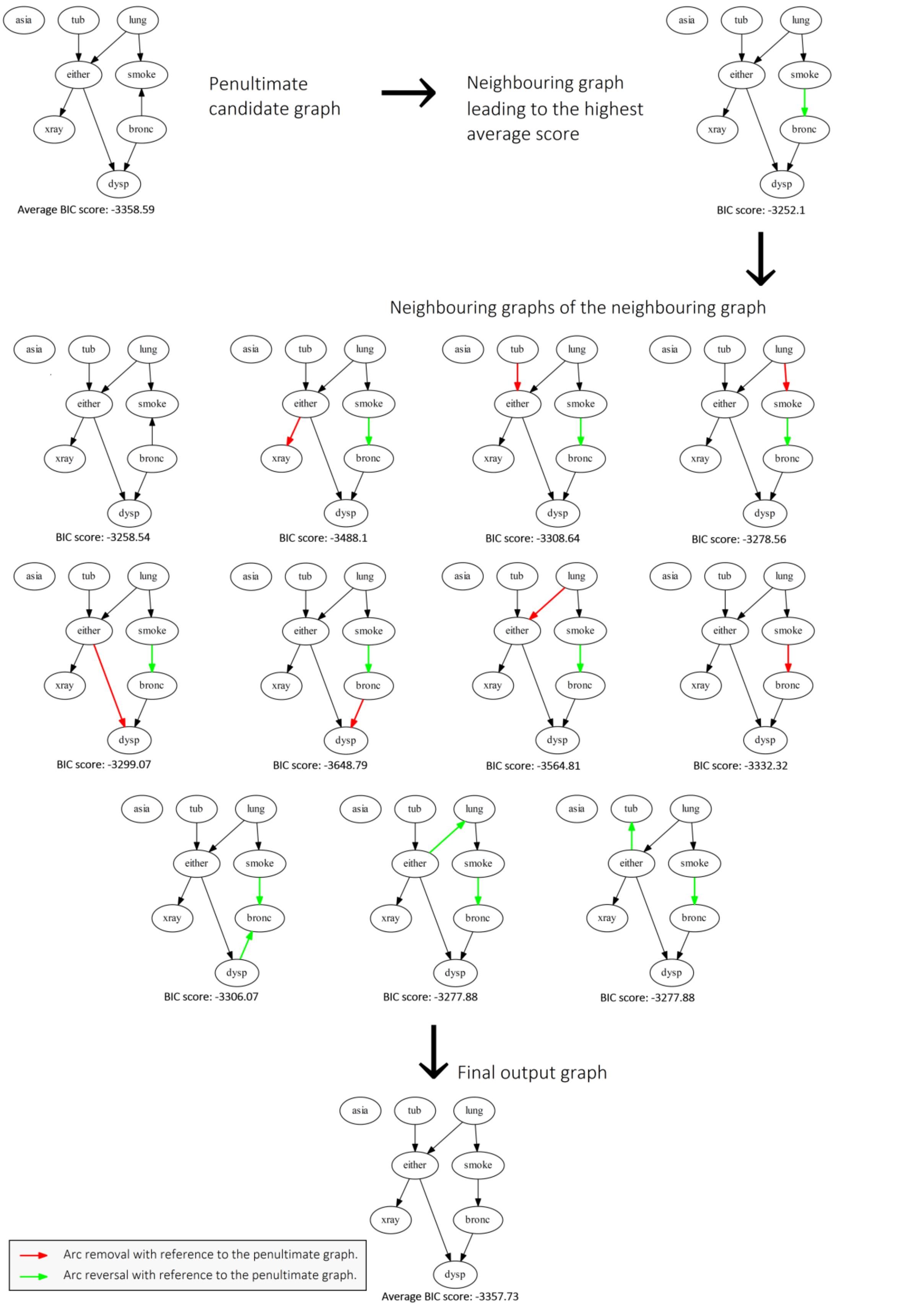

**Figure 2.** An illustration of the extended hill-climbing heuristic in MAHC, with application to the Asia network from subsection 2.1. The illustration focuses on the penultimate candidate graph and its neighbouring graph that maximises the average objective score. The BIC score of the final output graph represents the average BIC score across the 12 graphs; i.e., the neighbouring graph of the penultimate graph and its 11 neighbouring graphs.

**Algorithm 2:** The Model Averaging Hill-Climbing (MAHC) algorithm.

**Input:** data set $D$, max in-degree for pre-processing $M_P$, max in-degree for structure learning $M_S$, initial empty DAG $G$, BIC function given DAG $G$ and data $\boldsymbol{D}$.

**Output:** The DAG $G_{\max}$ that maximises the average set of scores $\overline{S(G,G_n)} = \overline{BIC(G,G_n,D)}$, where $G_n$ represents all the valid neighbours of $G$, in the search space of DAGs.

Apply **Algorithm 1** to data set $D$, to pre-process and prune CPSs up to in-degree $M_P$, and to obtain the set $F$ that contains the edges forbidden/pruned off.<br>
Obtain the score of the initial DAG $G$, $S_G = \overline{S(G,G_n)}$.<br>
Set $S_{\max} = S_G$ and $G_{\max} = G$.<br>
**while** $S_{\max}$ increases **do**<br>
    **for** every neighbouring graph $G_n$ of $G_{max}$ that satisfies acyclicity, the allowed arcs given $F$, and $M_S$, **do**<br>
        add $S(G_n)$ to a new set of objective scores $S_n$.<br>
            **for** every neighbouring graph $G_{nn}$ of $G$ that satisfies acyclicity, the allowed arcs given $F$, and $M_S$, **do**<br>
                add $S(G_{nn})$ to the set of objective scores $S_n$.<br>
            **end for**<br>
        get the score of DAG $G_n$, and set $S_{G_n} = \overline{S_n}$.<br>
        **if** $S_{G_n} > S_G$ **do** $S_G = S_{G_n}$ and $G = G_n$<br>
    **end for**<br>
    **if** $S_G > S_{\max}$, **do** $S_{\max} = S_G$ and $G_{\max} = G$<br>
**end while**

# 3. Evaluation, results and discussion

## *3.1. Evaluation process*

The evaluation process considers different algorithms, case studies, sample sizes, data assumptions and evaluation metrics. We cover these different settings in turn.

In terms of the **algorithms**, the MAHC algorithm is compared against the approximate score-based HC and TABU, and the exact score-based GOBNILP (Cussens, 2011) that relies on combinatorial optimisation and sound pruning to ensure that the highest possible graph is identified. We also consider the constraint-based PC-Stable (Spirtes and Glymour, 1991; Colombo and Maathuis, 2014) that assumes causal sufficiency, and its variant FCI (Spirtes et al., 1999) which does not assume causal sufficiency, and which might be more suitable for the noisy data experiments described below. Lastly, we also consider the hybrid MMHC (Tsamardinos et al., 2016) and SaiyanH (Constantinou, 2020) algorithms.

Because the MAHC algorithm described in this paper represents an extension of HC, it is important to validate MAHC with reference to the HC implementation on which MAHC is based. Since MAHC is implemented in Bayesys v3.1 (Constantinou, 2019), we used the Bayesys v3.1 software to test the HC, TABU and SaiyanH algorithms. Additionally, we used the GOBNILP v1 Python software (Cussens, 2011; Cussens and Bartlett, 2015) to test the GOBNILP algorithm, the rcausal v1.1.1 package that is part of the Tetrad software (Center for Causal Discovery, 2020) to test the PC-Stable and the FCI algorithms, and the bnlearn R v4.5 statistical package (Scutari, 2011) to test the MMHC algorithm. All algorithms are tested with their hyperparameter input defaults, which we provide in Table A1.

We consider the six **case studies** described in Table 2, taken from the Bayesys repository (Constantinou et al., 2020). The selected case studies offer a good range of old and new, as well as simple and complex, BNs that come from different application domains. They include three standard real-world networks, the Asia, Alarm and Pathfinder, and three recent real-world networks, the Sports, ForMed and Property, previously used in (Constantinou et al., 2021).

**Table 2.** The properties of the six case studies.

| Case study | Nodes | Arcs | Average degree | Maximum In-degree | Free parameters |
|---|---|---|---|---|---|
| Asia | 8 | 8 | 2.00 | 2 | 18 |
| Alarm | 37 | 46 | 2.49 | 4 | 509 |
| Pathfinder | 109 | 195 | 3.58 | 5 | 71,890 |
| Sports | 9 | 15 | 3.33 | 2 | 1,049 |
| ForMed | 88 | 138 | 3.14 | 6 | 912 |
| Property | 27 | 31 | 2.30 | 3 | 3,056 |

Regarding the different **data assumptions**, we consider noisy synthetic data, in addition to traditional clean synthetic data, generated for each of the six networks described above. The noisy synthetic data sets are made available by the Bayesys repository and come from the study by Constantinou et al (2021) who investigated the impact of different types of data noise on structure learning. The noisy data sets include missing values that are represented by a new state called 'missing' (this is to ensure that all algorithms can process the data), latent variables, measurement error, and reduced dimensionality in the form of merging two states into one when multinomial variables are present in the data. Each type of noise is randomised with a rate 5%. For example, each variable value has 5% chance to become 'missing'. This data set is denoted as cMISL in (Constantinou et al., 2021), and additional information regarding the construction of the data set can be found in that study. We denote the results obtained from traditional synthetic data as *Clean*, and the results obtained from noisy synthetic data as *Noisy*. We also consider the four sample sizes of $10^2$, $10^3$, $10^4$, and $10^5$ across all case studies.

We measure the accuracy of the learnt graphs with respect to the true graphs using five different graphical metrics. These are the Precision, Recall, F1, BSF and SHD. Precision $P$ and Recall $R$ range from 0 and 1 (lowest to highest score) and are measured by

$$P = \frac{TP}{TP + FP} \quad \text{and} \quad R = \frac{TP}{TP + FN}$$

where $TP$, $FP$ and $FN$ represent the number of *true positive*, *false positive*, and *false negative* edges in the learnt graph respectively. The F1 score also ranges from 0 to 1 and represents the harmonic mean between $P$ and $R$:

$$\text{F1} = 2\frac{P.R}{P + R}.$$

The BSF is a balanced score that ranges from -1 to 1, where -1 indicates that the learnt graph is the inverse of the true graph, 0 indicates that the learnt graph is as accurate as an empty or a fully connected graph, and 1 indicates a perfect match between the learnt and the true graphs (Constantinou, 2019b). The BSF score is obtained by

$$\mathrm{BSF} = \frac{1}{2}\left(\frac{\mathrm{TP}}{E_P} + \frac{\mathrm{TN}}{E_A} - \frac{\mathrm{FP}}{E_A} - \frac{\mathrm{FN}}{E_P}\right)$$

where $TN$ represents the number of *true negatives* in the learnt graph in terms of correctly determining the edges that are absent, and $E_P$ and $E_A$ represent the number of edges present and absent in the true graph respectively, where

$$E_A = \frac{|V| \times (|V| - 1)}{2} - E_P$$

Lastly, the SHD score represents the number of edge additions, deletions and reversals needed to convert the learnt graph into the true graph (Tsamardinos et al., 2016). We weight the penalty of an arc reversal by 0.5 relative to edge additions and deletions, to acknowledge the fact that an arc reversal implies correct discovery of an edge but with an incorrect direction. This assumption is applied to all five graphical metrics.

All the metrics are used to compare a) the graphs learnt with clean data in terms of recovering the true CPDAG, and b) the graphs learnt with noisy data in terms of recovering the true Partial Ancestral Graph (PAG). A PAG represents a set of Markov equivalence Maximal Ancestral Graphs (MAGs), and a MAG is an extension of DAG that can represent the presence of latent variables. We make this distinction because some of the variables are latent in the noisy data set. From the algorithms considered, FCI is the only one which does not assume causal sufficiency, and this means that it is the only algorithm that may produce bi-directed edges. Table 3 lists the penalty weights that are considered by all the graphical evaluation metrics covered above, in evaluating the learnt graphs with respect to the true CPDAGs or PAGs. As shown in Table 3, we penalise failure to predict bi-directed edges that may be present in the true PAGs derived from noisy experiments due to latent confounders (details in subsection 3.1). This means that FCI is the only algorithm that can, in theory, predict the true PAGs, whereas all other algorithms are guaranteed a penalty for each bi-directed edge present in each of those true PAGs.

In addition to the five graphical metrics described above, we consider the inference-based BIC score as defined by Equation 1. This model selection score enables us to investigate how the BIC scores of MAHC, which are driven by model averaging, compare with other score-based algorithms that employ the same BIC objective function as their model selection score.

Lastly, we measure computational complexity in terms of structure learning runtime. We apply a 6-hour runtime limit for all algorithms. If no result is returned within the runtime limit, we do not record a result. The same process is followed if an algorithm returns a runtime error. These details can be found in Table B1, which presents all the graphical and inference-based scores for all algorithms and over all the experiments.

**Table 3.** The penalty weights used by the graphical evaluation metrics when comparing the learnt CPDAGs or PAGs to the true CPDAG or true PAG.

| Rule | True graph | Learnt graph | Penalty | Reasoning |
|---|---|---|---|---|
| 1 | A → B | A → B | 0 | Complete match |
| 2 | A → B | A ↔ B, A − B , A ← B | 0.5 | Partial match |
| 3 | any edge | no edge | 1 | No match |
| 4 | A ↔ B | A ↔ B | 0 | Complete match |
| 5 | A ↔ B | A − B , A ← B, A → B | 0.5 | Partial match |
| 6 | A − B | A − B | 0 | Complete match |
| 7 | A − B | A ↔ B , A ← B, A → B | 0.5 | Partial match |
| 8 | no edge | no edge | 0 | Complete match |
| 9 | no edge | Any edge/arc | 1 | No match |

### *3.2. Graphical accuracy*

Direct comparisons between SHD scores involving networks of different size are inappropriate. For example, a SHD score of 5 might indicate poor performance for a network that contains 10 edges, and good performance for a network containing 100 edges. We resolve this bias by normalising the SHD score into a relative percentage value. To avoid normalisation issues caused by SHD scores of 0, we have increased all SHD scores, for all algorithms, by one. This was done only for the purposes of computing the relative percentages, which means that the SHD scores depicted in Table B.1 represent the raw scores prior to normalisation. For example, and as it can be seen in Table B1, HC generates an SHD score of 6.5 for the case of Asia for sample size $10^2$ assuming clean data, and this score is slightly higher (worse) than the corresponding SHD score of 5 generated by MAHC. This produces the score of 80% for HC relative to the score of MAHC (i.e., $\frac{(5+1)}{(6.5+1)}$), where a percentage score lower than 100% indicates worse performance relative to MAHC and a percentage score higher than 100% indicates better performance. On the other hand, at sample size $10^3$, both HC and MAHC generate an SHD score of 1, and this produces a relative score of 100% for HC which implies identical performance with MAHC.

Fig 3 provides a complete summary of graphical accuracy, where each chart corresponds to an evaluation metric. The charts on the left focus on clean data whereas the charts on the right on noisy data. For any given metric, the percentage scores represent the average relative difference in scores between the specified algorithm and MAHC, across all case studies and sample sizes. With reference to the example discussed in the previous paragraph, the average relative difference represents the average percentage score each algorithm generates relative to MAHC, for each of the metrics. That is, in the context of the two examples discussed above, the average relative difference of HC against MAHC would be the average over scores 80% and 100%; i.e., 90%. The negative percentage values depicted in Fig 3 indicate that the specified algorithm or learning class performed worse than MAHC (i.e., −10%, on average, with reference to the previous example), and these results are coloured in green and blue for algorithms and learning classes respectively, to reflect a positive outcome for MAHC. Similarly, positive percentage values indicate that the specified algorithm or learning class outperformed MAHC, and these results are coloured in red and orange for algorithms and learning classes respectively, to reflect a negative outcome for MAHC. Each learning class depicted at the bottom of each chart corresponds to the average scores of the two related algorithms, where constraint-based represents PC and FCI, score-based represents HC and TABU, and hybrid represents MMHC and SaiyanH. The GOBNILP algorithm itself can be viewed as a fourth class that relies on exact score-based learning that always identifies the highest scoring graph.

Because some algorithms did not produce a result for all the experiments (refer to Table B1), the results data include missing data points which may bias the conclusions. This is because missing results do not occur at random; they tend to occur on the more complex case studies, or when the sample size is big. Because of this, care should be given when reading the results in Fig 3, and particularly the results that involve PC-Stable and FCI, and to a much smaller extend GOBNILP, since these three algorithms did not complete all the experiments within the 6-hour runtime limit. With reference to PC-Stable and FCI, the comparisons illustrated in Fig 3 are largely based on simple and/or cases with limited sample size. A more complete picture about the relative performance of these algorithms is provided in the additional results that we discuss below.

Tables 4 and 5 summarise the graphical performance in terms of the percentage of times the score of MAHC was better, worse, or the same, across all experiments and relative to each of the other seven algorithms, for clean and noisy data respectively. In these tables, algorithms that fail to generate a score for a particular experiment are assumed to be ‘worse’ than algorithms than do generate a score. Because we consider algorithms spanning different classes of learning, we discuss the results of MAHC with reference to other algorithms distributed by learning class; i.e., a) approximate score-based, b) exact score-based, c) constraint-based, and d) hybrid. The discussion of the results that follows takes into consideration Fig 3 as well as both Tables 4 and 5.

**Graphical performance relative to approximate score-based learning:** Fig 3 suggests that MAHC outperforms HC, on average, across all five metrics and over both clean and noisy data. This is reiterated by the results in Tables 4 and 5 which show that, overall, HC generated a better score than MAHC in 22.5% of the clean data experiments, which decreases to 15% with noisy data. According to Fig 3, the discrepancy in scores with TABU is less obvious. Still, the overall results suggest that MAHC performs somewhat better than TABU. This result is repeated in Tables 4 and 5, where MAHC outperforms TABU in a higher number of experiments, and this difference increases in the presence of data noise. Overall, the results suggest that the graphical performance of MAHC is considerably better than the HC implementation on which MAHC is based, and somewhat better than TABU.

**Graphical performance relative to exact score-based learning:** As expected, because exact learning always returns the highest scoring graph, it outperforms all other algorithms in most cases. As shown in Tables 4 and 5, MAHC generates a better score than GOBNILP in 32.5% and a worse score in 50% of the clean experiments, but this reverses to 40% better scores and 38.3% worse scores under data noise. This result is consistent with the results against HC and TABU, that show that MAHC improves relative performance in the presence of data noise. As illustrated in Fig 3, while MAHC gains ground against GOBNILP under data noise, it outperforms GOBNILP only in average Precision.

**Graphical performance relative to constraint-based learning:** Comparisons against PC-Stable and FCI are less consistent and more uncertain, and this can be explained by the different learning class they represent as well as due to missing results. Fig 3 shows that constraint-based PC-Stable and FCI are generally better at Recall and Precision with clean data, but their relative performance decreases considerably under data noise, since MAHC gains a considerable advantage in Precision as well as reduces the disadvantage in Recall. While this result is repeated in terms of F1 and SHD scores, we observe the reverse conclusion in terms of BSF score. The overall results in Tables 4 and 5, however, show that MAHC generates a better score than PC-Stable and FCI in 50% and 57.5% of the clean experiments, and this increases to 66.7% and 60% of the noisy data cases. While the results against constraint-based learning are rather inconsistent, they do repeat the pattern from previous results where MAHC is found to perform better under data noise.

**Graphical performance relative to hybrid learning:** Hybrid learning MMHC and SaiyanH exhibit similar inconsistencies to constraint-based learning in their comparisons with MAHC. As shown in Tables 4 and 5, MMHC and SaiyanH generate better scores than MAHC in 22.5% and 45.8% of the clean experiments, and in 26.7% and 40.8% of the noisy experiments. Interestingly, MMHC is the only algorithm whose relative performance increases against MAHC in the presence of data noise. These patterns are somewhat repeated in by the relative average scores depicted in Fig 3, where MMHC performs poorly in F1 and BSF scores, and relatively well in SHD under noisy data, relative to MAHC. Still, the overall results show that MAHC outperforms MMHC across all BSF, F1 and SHD scores, with and without data noise. On the other hand, SaiyanH produced results that are more in line with MAHC, since they would sometimes be better or worse than those produced by MAHC, within the same and across different metrics. Overall, MAHC is found to be inferior to SaiyanH in clean experiments, but generally outperforms SaiyanH when the data are noisy, and this pattern is consistent with those obtained from comparisons against algorithms from the other classes of learning.

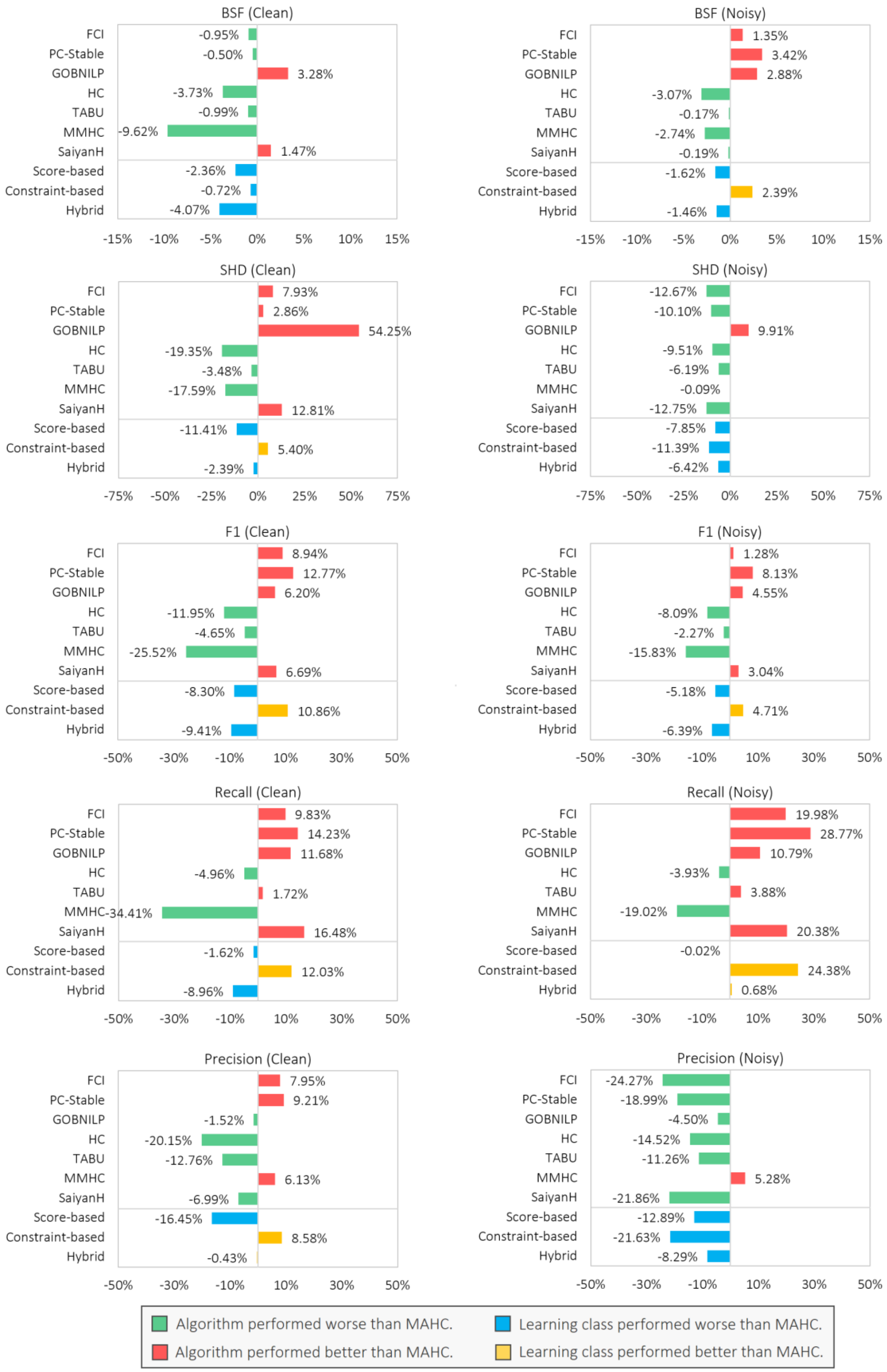

**Figure 3.** The graphical performance of the other seven algorithms, and four learning classes (including GOBNILP which represents the exact score-based learning class), relative to MAHC. A negative green/blue score implies the specified algorithm performed worse than MAHC, and a positive red/orange score better than MAHC. The results are separated by clean (graphs on the left) and noisy (graphs on the right) data.

**Table 4.** The graphical accuracy of MAHC relative to the other seven algorithms. The percentages represent the number of times the score of MAHC was better, worse or the same relative to each of the other algorithms, across all case studies and sample sizes, assuming *clean* data.

| | | BSF | | | SHD | | | F1 | | | Recall | | | Precision | | | Overall | | |
|---|---|---|---|---|---|---|---|---|---|---|---|---|---|---|---|---|---|---|---|
| Relative to: | Learning class | Better | Same | Worse | Better | Same | Worse | Better | Same | Worse | Better | Same | Worse | Better | Same | Worse | Better | Same | Worse |
| FCI | Constraint-based | 54.2% | 0.0% | 45.8% | 50.0% | 0.0% | 50.0% | 45.8% | 0.0% | 54.2% | 54.2% | 8.3% | 37.5% | 45.8% | 0.0% | 54.2% | 50.0% | 1.7% | 48.3% |
| PC-Stable | Constraint-based | 58.3% | 0.0% | 41.7% | 58.3% | 0.0% | 41.7% | 58.3% | 0.0% | 41.7% | 62.5% | 4.2% | 33.3% | 50.0% | 0.0% | 50.0% | 57.5% | 0.8% | 41.7% |
| GOBNILP | Exact score-based | 25.0% | 16.7% | 58.3% | 25.0% | 16.7% | 58.3% | 45.8% | 16.7% | 37.5% | 20.8% | 20.8% | 58.3% | 45.8% | 16.7% | 37.5% | 32.5% | 17.5% | 50.0% |
| HC | Score-based | 58.3% | 8.3% | 33.3% | 62.5% | 8.3% | 29.2% | 83.3% | 8.3% | 8.3% | 54.2% | 12.5% | 33.3% | 83.3% | 8.3% | 8.3% | 68.3% | 9.2% | 22.5% |
| TABU | Score-based | 29.2% | 16.7% | 54.2% | 37.5% | 16.7% | 45.8% | 58.3% | 20.8% | 20.8% | 29.2% | 16.7% | 54.2% | 66.7% | 16.7% | 16.7% | 44.2% | 17.5% | 38.3% |
| MMHC | Hybrid | 83.3% | 4.2% | 12.5% | 83.3% | 8.3% | 8.3% | 62.5% | 4.2% | 33.3% | 87.5% | 4.2% | 8.3% | 45.8% | 4.2% | 50.0% | 72.5% | 5.0% | 22.5% |
| SaiyanH | Hybrid | 37.5% | 12.5% | 50.0% | 33.3% | 8.3% | 58.3% | 54.2% | 12.5% | 33.3% | 29.2% | 8.3% | 62.5% | 66.7% | 8.3% | 25.0% | 44.2% | 10.0% | 45.8% |

**Table 5.** The graphical accuracy of MAHC relative to the other seven algorithms. The percentages represent the number of times the score of MAHC was better, worse, or the same relative to each of the other algorithms, across all case studies and sample sizes, assuming *noisy* data.

| | | BSF | | | SHD | | | F1 | | | Recall | | | Precision | | | Overall | | |
|---|---|---|---|---|---|---|---|---|---|---|---|---|---|---|---|---|---|---|---|
| Relative to: | Learning class | Better | Same | Worse | Better | Same | Worse | Better | Same | Worse | Better | Same | Worse | Better | Same | Worse | Better | Same | Worse |
| FCI | Constraint-based | 66.7% | 0.0% | 33.3% | 62.5% | 4.2% | 33.3% | 83.3% | 0.0% | 16.7% | 33.3% | 12.5% | 54.2% | 87.5% | 0.0% | 12.5% | 66.7% | 3.3% | 30.0% |
| PC-Stable | Constraint-based | 54.2% | 0.0% | 45.8% | 54.2% | 0.0% | 45.8% | 79.2% | 0.0% | 20.8% | 29.2% | 8.3% | 62.5% | 83.3% | 0.0% | 16.7% | 60.0% | 1.7% | 38.3% |
| GOBNILP | Exact score-based | 29.2% | 20.8% | 50.0% | 37.5% | 16.7% | 45.8% | 45.8% | 20.8% | 33.3% | 20.8% | 29.2% | 50.0% | 66.7% | 20.8% | 12.5% | 40.0% | 21.7% | 38.3% |
| HC | Score-based | 45.8% | 29.2% | 25.0% | 58.3% | 25.0% | 16.7% | 54.2% | 37.5% | 8.3% | 37.5% | 37.5% | 25.0% | 79.2% | 20.8% | 0.0% | 55.0% | 30.0% | 15.0% |
| TABU | Score-based | 33.3% | 20.8% | 45.8% | 41.7% | 20.8% | 37.5% | 50.0% | 25.0% | 25.0% | 20.8% | 33.3% | 45.8% | 75.0% | 16.7% | 8.3% | 44.2% | 23.3% | 32.5% |
| MMHC | Hybrid | 66.7% | 16.7% | 16.7% | 66.7% | 16.7% | 16.7% | 41.7% | 20.8% | 37.5% | 66.7% | 16.7% | 16.7% | 29.2% | 25.0% | 45.8% | 54.2% | 19.2% | 26.7% |
| SaiyanH | Hybrid | 41.7% | 4.2% | 54.2% | 50.0% | 0.0% | 50.0% | 79.2% | 4.2% | 16.7% | 25.0% | 4.2% | 70.8% | 83.3% | 4.2% | 12.5% | 55.8% | 3.3% | 40.8% |

### 3.3. *Model selection accuracy*

In this section, we investigate how the BIC score of the learnt graphs of MAHC (i.e., the BIC score of the output graph and not the average BIC score computed during structure learning) differs from the BIC score of the learnt graphs produced by the other algorithms. The constraint-based algorithms optimise the graph given a set of conditional independence results, so it may be unfair to compare them against algorithms that are designed to maximise BIC. Still, we decided to include all algorithms for reference, taking into account that constraint-based algorithms are not designed to maximise BIC.

Fig 4 presents the BIC score discrepancy each of the algorithms has relative to the BIC score generated by the ground truth graph. These results focus on clean data and are distributed by case study and sample size. Negative percentage values indicate lower (worse) BIC scores relative to the ground truth graph, whereas higher percentage values indicate higher (better) BIC scores. Fig 5 repeats the results of Fig 4, but for noisy data. In this case, the ground truth graph is converted into the corresponding Mixed Ancestral Graph (MAG) to account for the latent variables that form part of data noise. In computing the BIC scores, if the output of an algorithm is a CPDAG or a PAG, the graph is converted into a valid DAG or MAG, respectively, that is part of their corresponding equivalence class. In the context of this paper, the main differences between a MAG and a DAG are a) a directed edge in a MAG represents causal or ancestral relationship in the presence of latent variables, and b) a bi-directed edge represents an edge that does not exist in the true DAG and is the result of a latent confounder. MAGs may also include other type of edges that represent selection bias, but these are out of the scope of this paper. A detailed description of MAGs can be found in (Spirtes et al., 2000).

The BIC score is shown to be an asymptotically consistent scoring criterion for MAGs that are based on continuous data (Richardson and Spirtes, 2002), and has already been applied to Gaussian distributed data (Tsirlis et al., 2018; Rantanen et al. 2021). Currently, however, no suitable BIC function exists to compute the score of a MAG from discrete data. To obtain reasonably accurate BIC estimates of the ground truth MAGs, we decided to implement the rather basic approach that enumerates all possible DAGs that could have been produced via all directed combinations of bi-directed edges. We then considered the average BIC score obtained across all those DAG combinations to be the BIC score of the corresponding MAG. However, this approach can be impractical for large networks. As it can be seen in Fig 5, we did not include the BIC scores of the Pathfinder MAG, and this is because the ground truth MAG of Pathfinder produced 2.15 billion different DAGs and Directed Cyclic Graphs (DCGs), almost all of which had in-degree higher than 10. Because the Bayesys v3.1 package is restricted to a maximum in-degree of 8 (refer to Table A1), and because it was impractical to assess such a high number of relatively complex graphs, we had to skip this case study for the case of noisy data. However, as discussed below, excluding Pathfinder from the results is not expected to have an influence on the overall conclusions.

Further to what has been stated in Introduction, the hypothesis here is that MAHC is likely to produce graphs with worse BIC score compared to the scores produced by the other algorithms that maximise the BIC score, even if MAHC produces better graphical scores in subsection 3.2. This is because while the BIC score is used for model selection by the score-based and hybrid learning algorithms considered in this study, it is used for model averaging by MAHC. In other words, MAHC does *not* return the graph that maximises the BIC score, but rather a graph that maximises the average BIC score over a set of graphs.

Both Figs 4 and 5 show a clear pattern where many algorithms often produce higher BIC scores relative to the scores of the true graphs, and this is particularly evident when the sample size of the input data is low relative to the complexity of the network. This is not surprising since, like most other objective functions, the BIC score is known to be inadequate when the sample size of the input data is limited relative to the complexity of the network. This means that, in the presence of limited data, such an objective function might be incapable of identifying the ground truth graph as the highest, or one of the highest, scoring graphs. This, in turn, implies that the objective function will direct algorithms to graphical structures that overfit the network. In other words, if the objective function fails to score the ground truth graph highly, the structure learning algorithm will also fail to predict it.

In fact, the charts in Figs 4 and 5 show that it is still possible for the algorithms to produce a higher BIC score than the score of the true graph even when the sample size is sufficient. This supports the results in (Constantinou et al., 2021) who showed that a higher BIC score does not necessarily imply a more accurate causal graph, irrespective of the sample size of the data. As shown in Fig 5, this outcome is more strongly evidenced in the case of noisy data, and this is reasonable since the ground truth graph no longer reflects the true distributions of the input data. This, in turn, suggests that it might not be appropriate to focus entirely on the BIC score when judging the performance of structure learning algorithms on real cases, since real data inevitably incorporate different types of data noise.

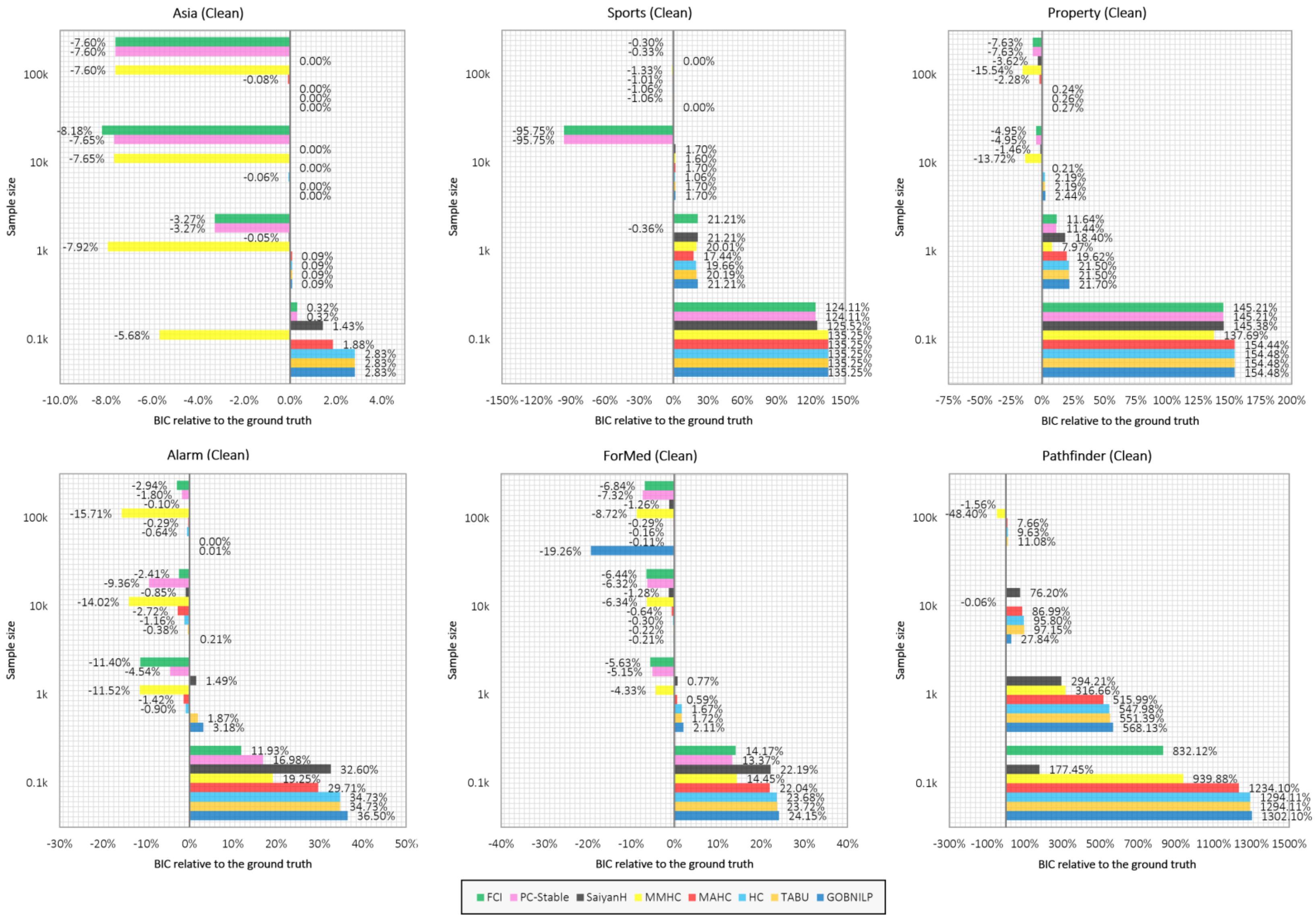

**Figure 4.** The BIC scores generated by the algorithms when applied to clean data, relative to the BIC scores of the true graphs, distributed per case study and sample size. Positive percentages indicate the learnt graphs produce a higher BIC score than the true graph, and vice versa for negative percentages.

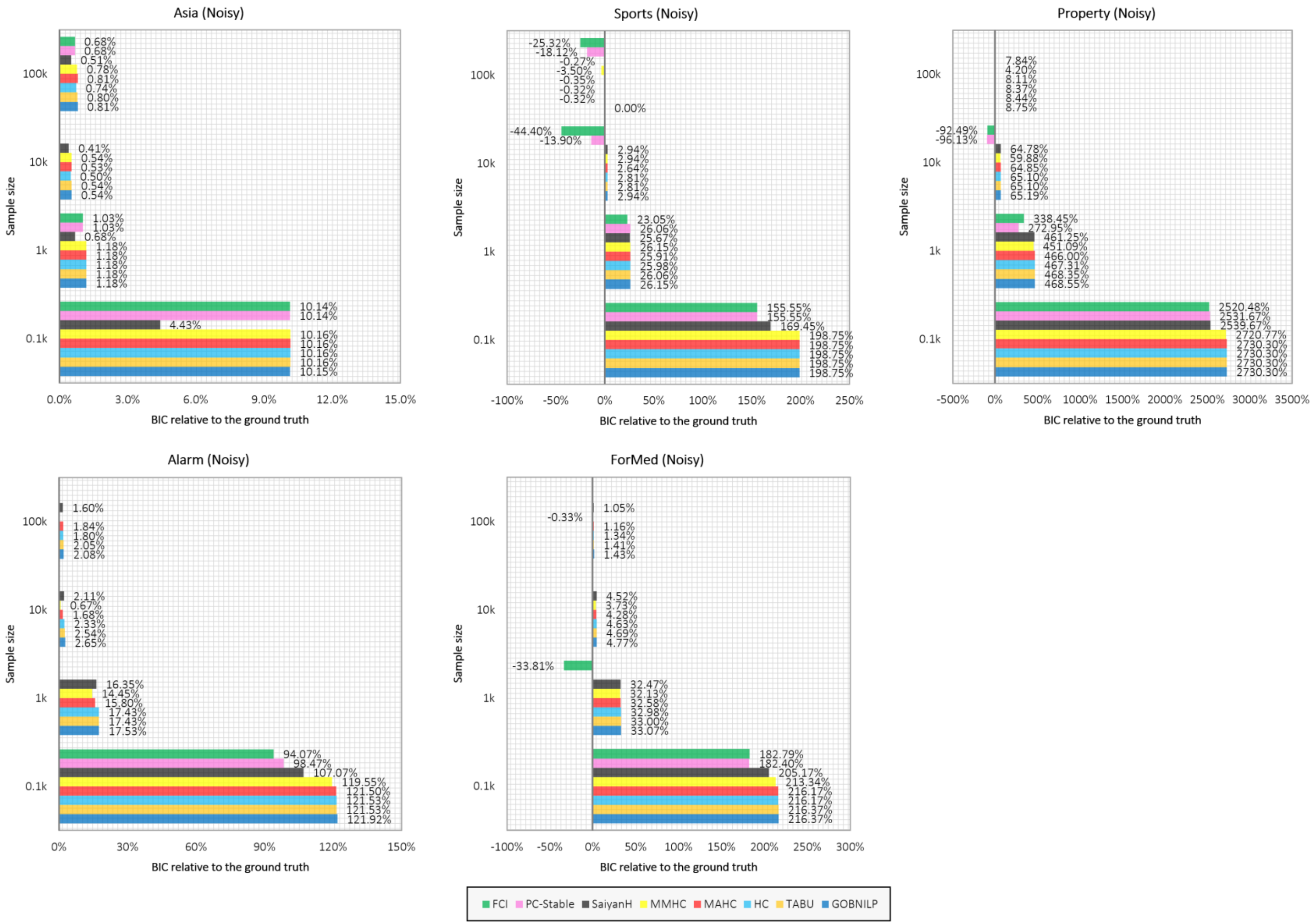

**Figure 5.** The BIC scores generated by the algorithms when applied to noisy data, relative to the BIC scores of the true graphs, distributed per case study and sample size. Positive percentages indicate the learnt graphs produce a higher BIC score than the true graph, and vice versa for negative percentages.

Table 6 summarises the BIC results in terms of average percentage score. This was achieved by normalising BIC scores between 0% (lowest/worse) and 100% (highest/best) in each experiment, having taken into consideration the scores produced by the algorithms as well as the scores produced by the ground truth graphs. For example, the average score of GOBNILP is 100% for clean data at $10^2$ sample size, because GOBNILP produced the highest score in all of the experiments that involved clean data with a sample size of $10^2$. For that same case, the average score of TABU is 99.16%, meaning the TABU score was, on average, 99.16% of the highest score in the experiments with sample size $10^2$.

Overall, the results show that GOBNILP will often produce graphs with the highest BIC score. Naturally, we would expect exact learning to produce the highest score every time, and the reason why this is not the case here is largely because of two known causes. Firstly, GOBNILP is restricted to a rather low maximum in-degree of 3, and this is necessary to achieve exact learning within reasonable computational times. However, this can be a problem on the basis that GOBNILP will never have the opportunity to score true graphs that contain in-degrees higher than 3, which is the case with half of the case studies considered. Under this scenario, GOBNILP might underperform relative to algorithms that do not impose such restriction and explore graphs that contain higher in-degrees. Secondly, GOBNILP failed to complete learning within the 6-hour runtime limit for some of the experiments, particularly in those where the sample size was higher, and those results reflect the accuracy of the graph obtained at the 6-hour runtime limit.

Other than GOBNILP, TABU is the algorithm that is most likely to generate the highest BIC score, closely followed by HC and MAHC. The hybrid algorithms which include a phase that relies on constraint-based learning, as well as the constraint-based algorithms themselves, generate considerably lower BIC scores as expected since they are not designed with the sole purpose to maximise BIC. The BIC scores of MAHC are much closer to those of HC and TABU, and considerably stronger than those produced by the hybrid and constraint-based learning algorithms. This outcome supports the initial hypothesis and it can be viewed as a positive result on the basis that the reduction in BIC score relative to HC, due to model averaging, appears to be small, especially in the presence of data noise, relative to the gain in graphical accuracy over HC, as illustrated in subsection 3.2.

When focusing on the algorithms, the overall results are not surprising. They show that exact score-based learning performs best (at least when GOBNILP completes learning), closely followed by score-based learning, followed by hybrid learning, and lastly by constraint-based learning which is not designed for this task in the same way score-based algorithms are. What could be classified as surprising, however, is the score of the ground truth graphs. The results show that the lower the sample size, the lower the chance the true graph will be part of the higher scoring graphs, as expected. However, this is true only for the clean data cases. The data noise drops the relative performance of the ground truth under all sample sizes, as expected, but also causes some unexpected imbalance in terms of how the scores differ between sample sizes. What might be surprising, however, is how low the score of the ground truth graph will often be relative to the scores produced by the other algorithms, and especially score-based algorithms. The only case where the ground truth graph generates the highest overall BIC scores is when the sample size is highest in the presence of clean data.

**Table 6.** The BIC performance of the algorithms and the ground truth graph. The percentages represent average normalised BIC scores, where a higher percentage indicates a better score. An average of 100% implies that the specified algorithm generated the highest BIC scores across *all* case studies, for each sample size and data type combination.

| Sample size | GOBNILP | TABU | HC | MAHC | MMHC | SaiyanH | PC-Stable | FCI | Ground truth |
|---|---|---|---|---|---|---|---|---|---|
| Clean data | | | | | | | | | |
| $10^2$ | 100% | 99.16% | 99.13% | 94.63% | 69.67% | 88.88% | 76.21% | 77.65% | 11.44% |
| $10^3$ | 100% | 97.08% | 93.77% | 88.19% | 40.68% | 91.19% | 35.15% | 43.91% | 42.70% |
| $10^4$ | 90.21% | 98.66% | 97.42% | 92.50% | 18.14% | 90.22% | 20.77% | 28.45% | 80.96% |
| $10^5$ | 80.00% | 86.66% | 85.84% | 83.94% | 9.99% | 93.56% | 57.27% | 56.94% | 98.15% |
| Noisy data | | | | | | | | | |
| $10^2$ | 99.98% | 99.97% | 99.97% | 99.96% | 99.73% | 86.53% | 95.16% | 94.76% | 00.00% |
| $10^3$ | 100% | 99.92% | 99.86% | 99.54% | 99.32% | 91.06% | 69.09% | 66.42% | 30.81% |
| $10^4$ | 100% | 99.68% | 98.89% | 97.83% | 95.70% | 97.45% | 19.45% | 16.54% | 72.34% |
| $10^5$ | 100% | 98.69% | 94.98% | 94.37% | 47.22% | 84.24% | 59.25% | 41.89% | 34.51% |

### *3.4. Computational complexity*

The computational complexity of MAHC is measured empirically in terms of structure learning runtime. Fig 6a illustrates the cumulative structure learning runtime of each algorithm over all case studies and sample sizes, starting from the clean data experiments and moving to the noisy experiments. Fig 6b orders the algorithms by total structure learning runtime, separated into clean and noisy data experiments. The results are produced using different CPUs, and runtime is adjusted relative to each CPU's single-core benchmark performance[7]. The results assume a minimum of 1 second runtime per experiment. Moreover, algorithms that fail to complete learning within the 6-hour runtime limit are assigned a 6-hour runtime for each of those experiments. This applies to GOBNILP, PC-Stable and FCI algorithms, and implies that the true structure learning runtime of these algorithms could be much higher than that shown in Fig 6. Note that runtime comparisons between algorithms are not only dependent on the algorithm, but also on the quality of the code as well as the programming language. MAHC, HC, TABU and SaiyanH[8] are all part of the same package, which means that runtime differences between them are almost entirely explained by algorithmic, and not implementation, differences. On the other hand, the other algorithms come from different packages, which means that part of the runtime difference might be due to code quality or programming language. This is not necessarily a limitation, since the advantage of this approach is that comparisons reflect real-world, rather than theoretical, performance.

The results show that MAHC is slightly slower than HC and significantly faster than many of the other algorithms. This result is very encouraging given that model averaging represents a notoriously expensive computational process. It is worth mentioning that we first implemented MAHC without pruning. In the absence of pruning, MAHC was the slowest algorithm, by far; essentially model averaging rendered the first attempt of the algorithm impractical. Therefore, the computational efficiency of MAHC can be fully attributed to the pruning strategy described in subsection 2.1.

[7] Algorithms HC, TABU, SaiyanH, and MAHC were tested on a Ryzen 9 5950X, algorithms PC-Stable and GOBNILP on an Intel i5 5350U, algorithm FCI on an Intel i5 5350U and Intel i5 7360U, and algorithm MMHC on an Intel i5 8250 and an Intel i5 7360U. Each CPU's single-core performance was adjusted relative to the base speed of Ryzen 3 3600, as in (Constantinou et al., 2021), on the basis it had the highest market share. The adjustments involved *increasing* the runtime of HC, TABU, SaiyanH and MAHC by 23.23%, and *decreasing* the runtime of GOBNILP and PC-Stable by 41.69%, and FCI and MMHC by 24%.

[8] The efficiency of SaiyanH shown in Fig 6 is noticeably better compared to what is shown in (Constantinou et al., 2021), and this is due to implementation improvements and bug fixes in Bayesys v3.1, compared to the Bayesys v1.36 used in (Constantinou et al., 2021).

Other results of interest include those from GOBNILP. We took advantage of GOBNILP's option to retrieve the best graph discovered at the end of the 6-hour time limit. In two of the experiments (refer to Table B1), however, it was not possible to retrieve a graph since GOBNILP did not complete the pruning/pre-processing phase within the 6-hour time. This means that the overall runtime of GOBNILP depicted in Fig 6 considerably underestimates its true learning runtime needed to complete learning. Moreover, unlike most of the other algorithms, GOBNILP was considerably more efficient given noisy data. This outcome might indicate an issue with exact learning, in that the sound pruning strategy employed in GOBNILP might not remain *sound* in the presence of data noise, since it is possible that the improved efficiency is the result of pruning a higher number of CPSs due to data noise. This outcome is also observed for MAHC that also relies on a relevant pruning strategy, and whose runtime has also decreased in the presence of data noise. As shown in Fig 7, the highest rates of pruning for MAHC occurred in the presence of noisy data.

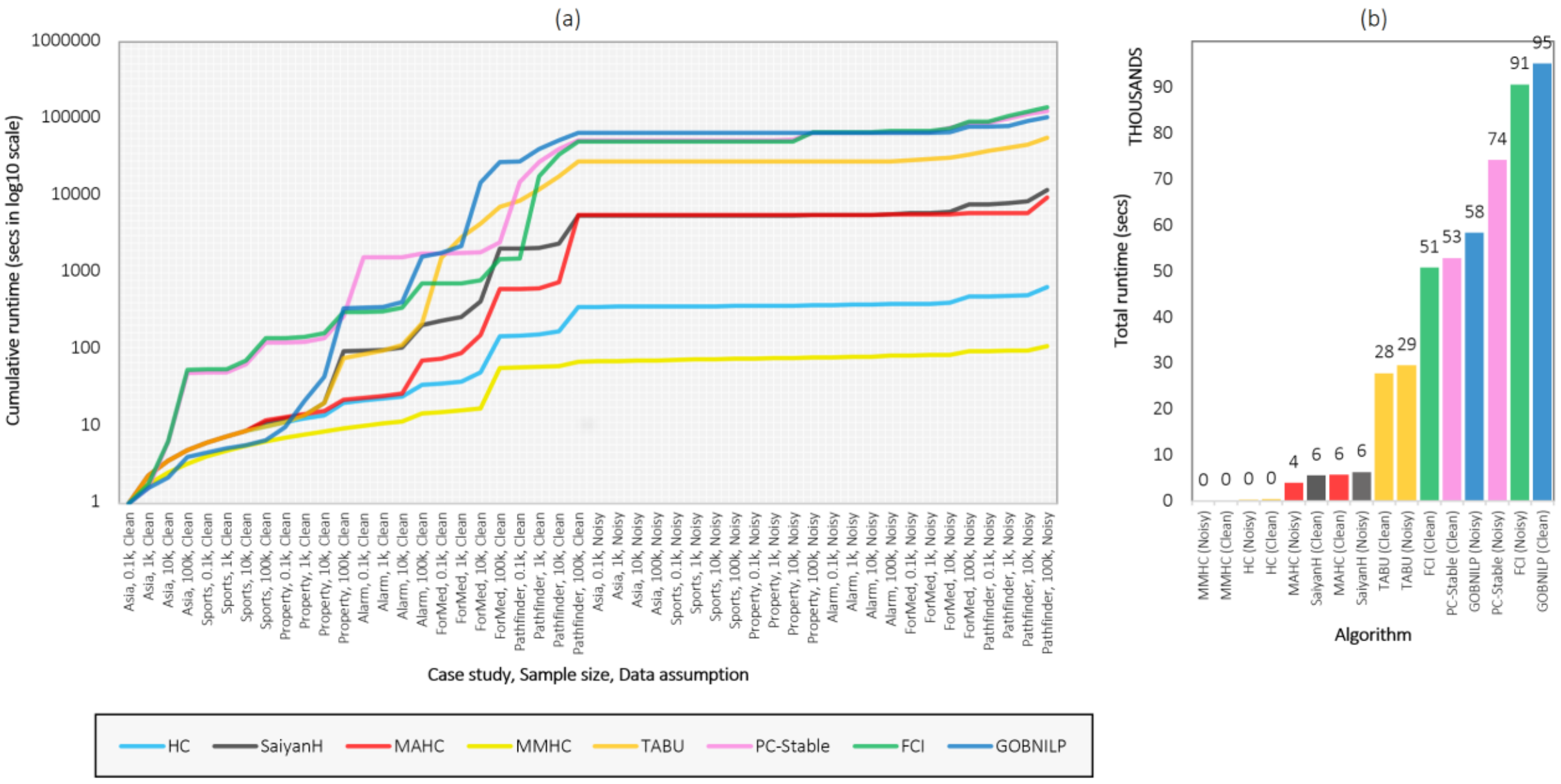

**Figure 6.** a) The cumulative structure learning runtime generated by each algorithm over each case study, sample size, and over both clean and noisy data sets. Cumulative lines in orange colour are measured with respect to the primary axis, and lines in black colour are measured with respect to the secondary axis. b) The total structure learning runtime of each algorithm over all experiments, separated by clean and noisy data.

Lastly, Fig 7 illustrates the percentage of edges that are pruned off during pre-processing, where pruning rates are separated by the level of in-degree. For example, pruning that occurs at in-degree 1 refers to the edges pruned off when comparing the local scores between empty CPSs and CPSs of size 1 that involve a single parent. Similarly, pruning that occurs at in-degree 2 refers to the edges pruned off when comparing the local scores obtained at CPSs of size 2 with the scores obtained at CPSs of lower size. As shown in Fig 7, it is much more likely that pruning decisions will occur early in the pre-processing process. From computational complexity point of view, this observation is important because it implies that considerably fewer CPSs of size 2 and 3 will be visited, which take longer to pre-process. It also means that increasing the maximum in-degree during the pre-processing step can be relatively efficient.

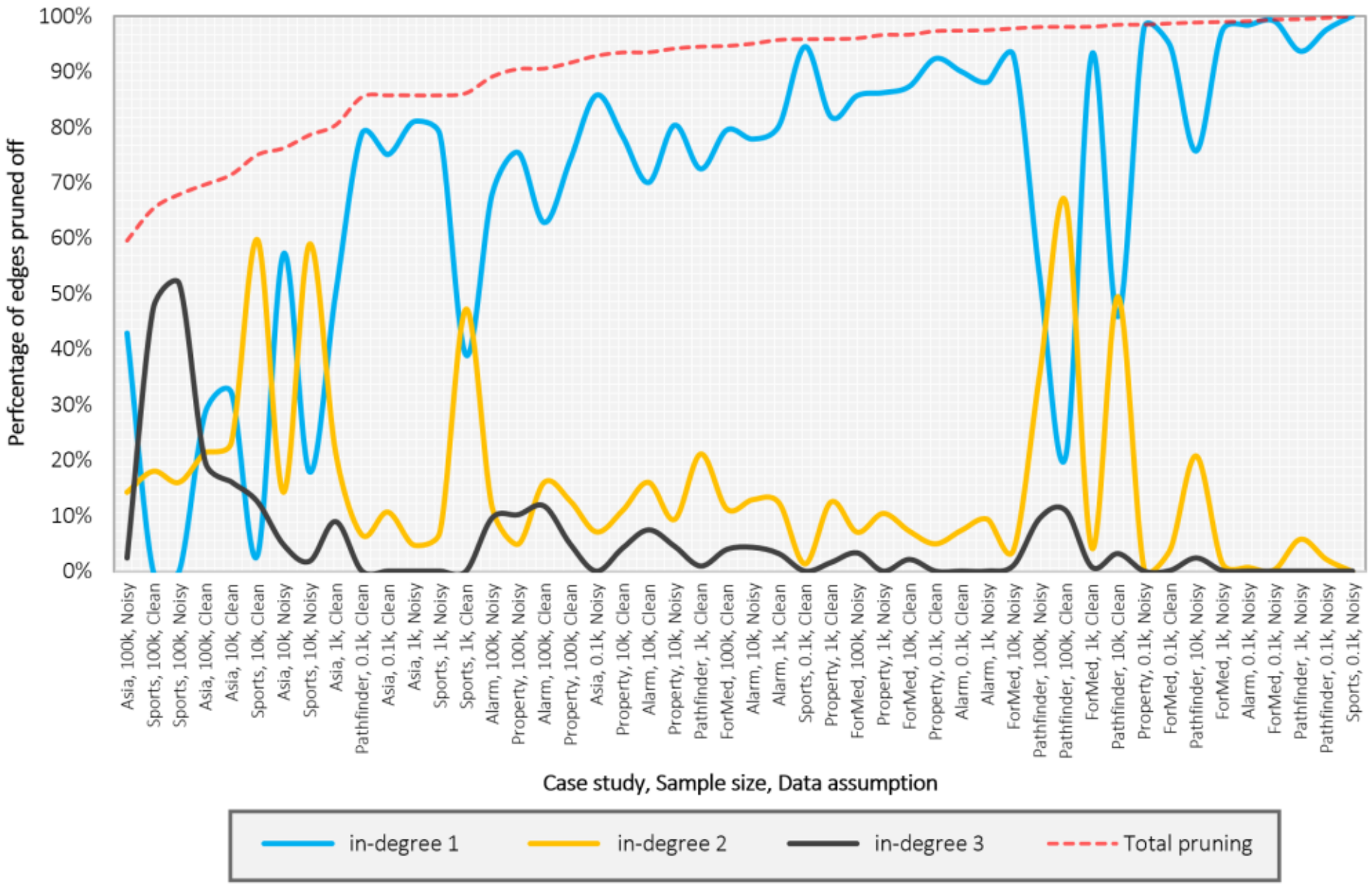

**Figure 7.** The percentage of edges pruned off at each level of in-degree during pre-processing. The case studies are ordered by lowest total pruning.

### *3.5. Assessing each learning phase of MAHC independently*

The previous subsections focused on how MAHC performs with both innovations; i.e., when pruning is combined with model averaging. In this subsection, we explore how each of these two learning phases influence hill-climbing independently. Specifically, we investigate how MAHC behaves:

- with pruning only; i.e., when we exclude the model averaging phase,
- with model averaging only; i.e., when we exclude the pruning phase,

and we compare those two independent phases against:

- HC; i.e., when we exclude both the pruning and model averaging phases,
- MAHC; i.e., when we include both the pruning and model averaging phases.

Fig 8 presents the results obtained across these comparisons, in terms of graphical scores, model selection scores, number of free parameters, edges learnt, and runtime, over each case study, sample size, and over both the clean and noisy data sets described in subsection 3.1. However, note that the comparisons with *MAHC_onlyAveraging* exclude 10 experiments that fail to complete structure learning within 12 hours. This is because *MAHC_onlyAveraging* performs model averaging in the absence of pruning, and so efficiency is a problem. The 10 experiments skipped for comparisons involving *MAHC_onlyAveraging* involve the ForMed and Pathfinder case studies at the highest sample sizes, over both clean and noisy data sets.

Table 7 summarises the results of Fig 8. Note that the scores depicted in Table 7 represent the difference between averaged scores, and this difference is not directly comparable to the scores depicted in Fig 3 which represent the average relative differences over each experiment, as described in subsection 3.2. Overall, the results show that, independently, each learning phase of MAHC has a positive effect on HC in terms of graphical accuracy, a negative effect on time complexity, and a mixed – although largely negative - effect on BIC score. Results of interest include:

a) Both pruning and model averaging phases have positive effects on graphical accuracy. When both phases are combined into the MAHC algorithm, their positive effect is in most cases stronger than their additive independent effects. This is an interesting observation and applies to both clean and noisy data.

b) The pruning phase reduces the number of free parameters and edges as one would expect, but this tends to also decrease the BIC scores, especially because the pruning is not sound. Interestingly, the model averaging phase also decreases dimensionality, but has a positive effect on BIC. When both phases are combined into the MAHC algorithm, their effect on dimensionality is often stronger than additive, whereas their effect on BIC score hovers between the BIC scores produced by each phase independently, and this translates into a negative overall impact on BIC.

c) Both learning phases increase runtime considerably. The pruning phase increases runtime because it needs to pre-process a large number of CPSs that are never visited by HC in the absence of pruning, whereas the model averaging phase increases runtime by a much larger amount, as one would expect. Note that the increase in runtime under the model averaging phase does not include the 10 cases which did not complete learning within the first 12 hours, which means that the true negative impact on runtime is much greater in the case of the model averaging phase. Interestingly, when both phases are combined into MAHC, the increase in runtime is lower compared to those produced by each learning phase independently in the case of clean data, but not in the case of noisy data. This can be explained by the results in Fig 8 which show that the complete MAHC algorithm often produces considerably fewer edges than HC.

**Table 7.** Assessment of the pruning and model averaging phases of MAHC, relative to HC.

| Algorithm (or learning phase) | BSF | F1 | Recall | Precision | SHD | BIC | Free param | Edges learnt | Runtime per case (secs) |
|---|---|---|---|---|---|---|---|---|---|
| | | | | Clean data | | | | | |
| MAHC_onlyPruning | 4.27% | 6.97% | -0.05% | 16.21% | -14.06% | -1.77% | -24.55% | -16.59% | 294 |
| MAHC_onlyAveraging | 1.12% | 4.35% | 3.03% | 6.60% | -3.61% | 0.19% | -14.58% | -3.33% | 3,021 |
| MAHC | 13.85% | 17.46% | 8.64% | 28.17% | -20.84% | -1.03% | -23.72% | -19.24% | 170 |
| | | | | Noisy data | | | | | |
| MAHC_onlyPruning | 6.69% | 3.75% | -0.48% | 9.86% | -4.84% | -0.46% | -21.95% | -13.22% | 64 |
| MAHC_onlyAveraging | 16.90% | 2.86% | 6.93% | 3.38% | -0.14% | 0.12% | -2.45% | -0.91% | 3,879 |
| MAHC | 16.97% | 12.44% | 8.24% | 17.88% | -7.17% | -0.31% | -26.69% | -13.83% | 118 |
| | | | | Overall | | | | | |
| MAHC_onlyPruning | 5.19% | 5.62% | -0.22% | 13.22% | -9.27% | -0.94% | -23.07% | -15.24% | 179 |
| MAHC_onlyAveraging | 7.18% | 3.71% | 4.62% | 5.04% | -1.45% | 0.15% | -6.80% | -2.24% | 2,786 |
| MAHC | 15.03% | 15.36% | 8.48% | 23.33% | -13.74% | -0.57% | -25.42% | -17.08% | 144 |

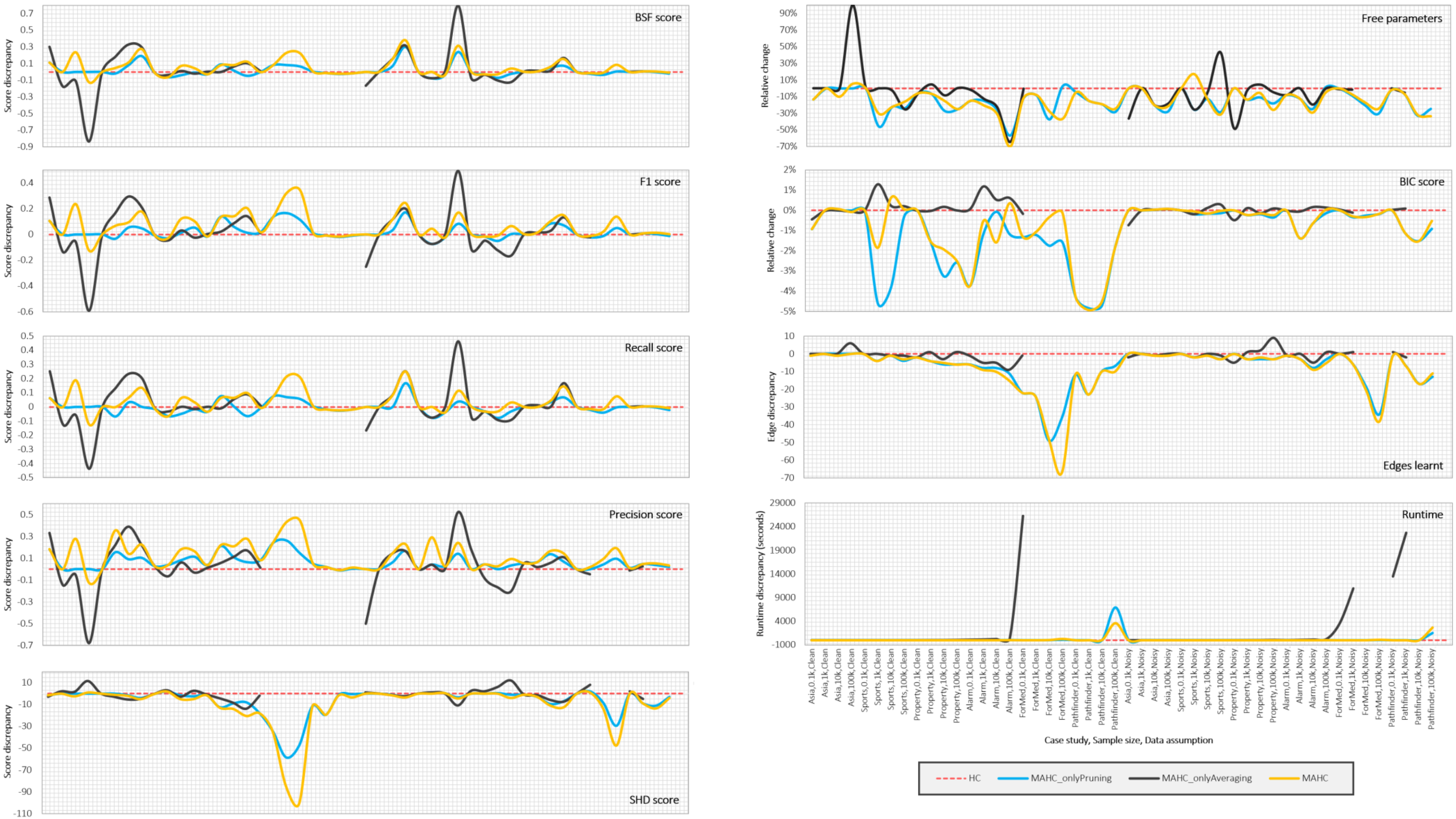

**Figure 8.** Comparisons between HC, MAHC_onlyPruning (excludes the model averaging phase), MAHC_onlyAveraging (excludes the pruning phase), and MAHC in terms of graphical scores, model selection scores, number of free parameters, edges learnt, and runtime, over each case study, sample size, and over both clean and noisy data sets.

# 4. Concluding remarks

This paper described a novel BN structure learning algorithm, called MAHC, that combines model averaging with pruning and HC heuristic search. The contribution can be viewed as twofold on the basis that both the model averaging and pruning phases of the algorithm are novel, and there is no reason to assume why they could not be applied to other suitable algorithms independently. Each learning phase is assessed independently and jointly. The results show that each learning phase has a positive effect on the graphical scores of HC, and remarkably, that their combined positive effect is often stronger than their additive independent effects.

The results, in terms of recovering the true graphical structure (refer to subsection 3.2), suggest that the performance of MAHC is competitive when the input data are clean, and often superior when the input data are noisy. The results suggest that model averaging strategies could be better suited for learning from real data, under the assumption that real observations never satisfy the ideal conditions assumed in clean synthetic experiments and that they often incorporate different kinds of data noise, many of which might be similar to those assumed in this paper. In terms of computational complexity, the empirical results show that MAHC is a few times slower than the very efficient HC and MMHC, but many times faster than algorithms such as PC-Stable, FCI and GOBNILP.

Additionally, the paper highlights some contradictory evidence regarding exact learning. In particular, the results show that the ground truth graph will not only not have the highest objective score, but will also often deviate considerably from the highest scoring graph, either due to data noise or limited data, both of which distort model fitting. This decreases the incentive to search for the highest scoring graph, and at the same time increases the importance of approximate learning. Moreover, the results also show that while the model averaging process in MAHC has caused the algorithm to recover graphs with slightly lower BIC score, these lower scoring graphs were often considerably more accurate in terms of graphical metrics (e.g., in terms of F1, BSF, and SHD scores).

On the other hand, the exact GOBNILP algorithm was found to be superior to all other algorithms, in general, both in terms of BIC and graphical metrics as well as both with clean and noisy data, at the expense of considerably higher computational complexity that becomes prohibitive for larger networks. MAHC did, however, close the gap in performance relative to GOBNILP in noisy data experiments. While the advantage gained in the presence of data noise did not help MAHC outperform GOBNILP in terms of BSF, F1 and SHD scores, it did help MAHC to marginally outperform GOBNILP in terms of the percentage of times a score is higher or lower than those generated by GOBNILP (refer to Table 5).

Future work in pruning and model averaging for approximate BN structure learning is exciting, mainly because, with the exception of sound pruning for exact structure learning, these approaches remain largely under-investigated. MAHC provides a set of interesting strategies that can be extended or combined with other methods to improve structure learning. For example, while it is reasonable to assume that what has been demonstrated in this paper extends to objective functions other than BIC score and search algorithms other than HC, this is something that needs to be further investigated. Lastly, the way MAHC operates on neighbouring graphs means that its computational efficiency can be further improved by utilising more than one CPU core (the implementation tested in this paper makes use of a single core). Future work will explore this direction, where one possible approach involves dividing neighbouring graphs by the number of available processing threads, thereby maximising the potential of parallel processing.

# **Appendix A:** Hyperparameter inputs

**Table A.1.** The default hyperparameter inputs for each algorithm, where $a$ is the dependency threshold [with dependency score], MID is the maximum in-degree, $iss$ is the imaginary sample size (also known as equivalent sample size) for BDeu, and $|V|$ is the size of variable set $V$. Learning classes C, S, ES, and H represent constraint-based, score-based, exact score-based, and hybrid learning respectively.

| Algorithm | Learning class | Software | Default hyperparameters |
|---|---|---|---|
| PC-Stable | C | Tetrad/rcausal v1.1.1 | $a = 0.01[G^2]$ |
| FCI | C | Tetrad/rcausal v1.1.1 | $a = 0.01[Chi^2]$ |
| MMHC | H | bnlean v4.5 | $a = 0.05[\text{MI}]$, score BIC |
| HC | S | Bayesys v3.1 | $\text{MID} = 8$, score BIC |
| TABU | S | Bayesys v3.1 | escapes $|V|(|V| - 1)$, $\text{MID} = 8$, score BIC |
| SaiyanH | H | Bayesys v3.1 | $a = 0.05[\text{MMD}]$, escapes $|V|(|V| - 1)$, $\text{MID} = 8$, score BIC |
| MAHC | S | Bayesys v3.1 | $\text{MID}(\text{pruning}) = 3$, $\text{MID}(\text{learning}) = 8$, score BIC |
| GOBNILP | ES | GOBNILP/pygobnilp v1 | $\text{MID} = 3$, score BIC |

# **Appendix B:** Raw performance scores

**Table B.1.** Raw scores generated by each of the algorithms for each of the experiments. Empty cells indicate a non-applicable score or failure to obtain a result within the 6-hour structure learning runtime limit.

| Algorithm | Case study | Sample size | Data | $BIC_{log2}$ | CPDAG scores BSF | SHD | F1 | Recall | Precision |
|---|---|---|---|---|---|---|---|---|---|
| MAHC | Asia | $10^2$ | Clean | -384.1 | 0.375 | 5 | 0.462 | 0.375 | 0.6 |
| MAHC | Asia | $10^3$ | Clean | -3252.1 | 0.875 | 1 | 0.933 | 0.875 | 1 |
| MAHC | Asia | $10^4$ | Clean | -32468.5 | 1 | 0 | 1 | 1 | 1 |
| MAHC | Asia | $10^5$ | Clean | -322966.6 | 0.875 | 1 | 0.875 | 0.875 | 0.875 |
| MAHC | Sports | $10^2$ | Clean | -1952.8 | 0.067 | 14 | 0.118 | 0.067 | 0.5 |
| MAHC | Sports | $10^3$ | Clean | -17057.4 | 0.267 | 11 | 0.4 | 0.267 | 0.8 |
| MAHC | Sports | $10^4$ | Clean | -157653.0 | 0.319 | 10.5 | 0.44 | 0.367 | 0.55 |
| MAHC | Sports | $10^5$ | Clean | -1565703.7 | 0.324 | 11 | 0.5 | 0.467 | 0.538 |
| MAHC | Property | $10^2$ | Clean | -5173.4 | 0.203 | 26.5 | 0.289 | 0.21 | 0.464 |
| MAHC | Property | $10^3$ | Clean | -40578.0 | 0.494 | 17.5 | 0.596 | 0.5 | 0.738 |
| MAHC | Property | $10^4$ | Clean | -354811.3 | 0.716 | 11.5 | 0.763 | 0.726 | 0.804 |
| MAHC | Property | $10^5$ | Clean | -3462242.5 | 0.781 | 9.5 | 0.803 | 0.79 | 0.817 |
| MAHC | Alarm | $10^2$ | Clean | -2344.0 | 0.22 | 40.5 | 0.3 | 0.228 | 0.438 |
| MAHC | Alarm | $10^3$ | Clean | -17761.0 | 0.64 | 17.5 | 0.711 | 0.641 | 0.797 |
| MAHC | Alarm | $10^4$ | Clean | -158185.7 | 0.79 | 11.5 | 0.811 | 0.793 | 0.83 |
| MAHC | Alarm | $10^5$ | Clean | -1512196.9 | 0.865 | 9 | 0.86 | 0.87 | 0.851 |
| MAHC | ForMed | $10^2$ | Clean | -7089.6 | 0.2 | 132.5 | 0.271 | 0.207 | 0.396 |
| MAHC | ForMed | $10^3$ | Clean | -64112.6 | 0.502 | 73.5 | 0.604 | 0.504 | 0.755 |
| MAHC | ForMed | $10^4$ | Clean | -610191.1 | 0.72 | 40.5 | 0.783 | 0.721 | 0.858 |
| MAHC | ForMed | $10^5$ | Clean | -6021128.5 | 0.846 | 26 | 0.87 | 0.848 | 0.893 |
| MAHC | Pathfinder | $10^2$ | Clean | -6661.5 | 0.084 | 209.5 | 0.135 | 0.09 | 0.273 |
| MAHC | Pathfinder | $10^3$ | Clean | -53304.1 | 0.13 | 228.5 | 0.184 | 0.141 | 0.264 |
| MAHC | Pathfinder | $10^4$ | Clean | -433290.6 | 0.223 | 238.5 | 0.27 | 0.238 | 0.31 |
| MAHC | Pathfinder | $10^5$ | Clean | -3696840.9 | 0.462 | 157 | 0.524 | 0.472 | 0.59 |
| MAHC | Asia | $10^2$ | Noisy | -663.5 | 0.167 | 5 | 0.25 | 0.167 | 0.5 |
| MAHC | Asia | $10^3$ | Noisy | -5921.8 | 0.333 | 4 | 0.444 | 0.333 | 0.667 |
| MAHC | Asia | $10^4$ | Noisy | -58390.3 | 0.367 | 5 | 0.5 | 0.5 | 0.5 |
| MAHC | Asia | $10^5$ | Noisy | -575873.9 | 0.55 | 4.5 | 0.643 | 0.75 | 0.563 |
| MAHC | Sports | $10^2$ | Noisy | -1924.8 | 0 | 13 | 0 | 0 | 0 |
| MAHC | Sports | $10^3$ | Noisy | -17854.6 | 0.269 | 9.5 | 0.412 | 0.269 | 0.875 |
| MAHC | Sports | $10^4$ | Noisy | -168376.0 | 0.385 | 8 | 0.5 | 0.385 | 0.714 |
| MAHC | Sports | $10^5$ | Noisy | -1661639.6 | 0.29 | 9.5 | 0.478 | 0.423 | 0.55 |
| MAHC | Property | $10^2$ | Noisy | -6133.0 | 0.125 | 28 | 0.205 | 0.125 | 0.571 |
| MAHC | Property | $10^3$ | Noisy | -52752.2 | 0.281 | 23 | 0.383 | 0.281 | 0.6 |
| MAHC | Property | $10^4$ | Noisy | -485062.2 | 0.637 | 12.5 | 0.732 | 0.641 | 0.854 |
| MAHC | Property | $10^5$ | Noisy | -4711486.7 | 0.795 | 11 | 0.8 | 0.813 | 0.788 |
| MAHC | Alarm | $10^2$ | Noisy | -4222.5 | 0.061 | 45 | 0.115 | 0.067 | 0.429 |
| MAHC | Alarm | $10^3$ | Noisy | -36679.1 | 0.263 | 35 | 0.358 | 0.267 | 0.545 |
| MAHC | Alarm | $10^4$ | Noisy | -340702.5 | 0.569 | 24 | 0.619 | 0.578 | 0.667 |
| MAHC | Alarm | $10^5$ | Noisy | -3296616.9 | 0.653 | 34 | 0.602 | 0.689 | 0.534 |
| MAHC | ForMed | $10^2$ | Noisy | -10656.8 | 0.082 | 130.5 | 0.14 | 0.082 | 0.479 |
| MAHC | ForMed | $10^3$ | Noisy | -98411.6 | 0.173 | 123.5 | 0.261 | 0.175 | 0.51 |
| MAHC | ForMed | $10^4$ | Noisy | -951093.4 | 0.366 | 105 | 0.454 | 0.371 | 0.584 |
| MAHC | ForMed | $10^5$ | Noisy | -9375124.2 | 0.56 | 99 | 0.584 | 0.571 | 0.597 |
| MAHC | Pathfinder | $10^2$ | Noisy | -12546.1 | 0.044 | 228.5 | 0.081 | 0.046 | 0.362 |
| MAHC | Pathfinder | $10^3$ | Noisy | -112457.1 | 0.038 | 250 | 0.072 | 0.043 | 0.217 |
| MAHC | Pathfinder | $10^4$ | Noisy | -1038565.9 | 0.122 | 253.5 | 0.188 | 0.133 | 0.321 |

| Algorithm | Case study | Sample size | Data | $BIC_{log2}$ | CPDAG scores BSF | SHD | F1 | Recall | Precision |
|---|---|---|---|---|---|---|---|---|---|
| MAHC | Pathfinder | $10^5$ | Noisy | -9693237.8 | 0.348 | 190 | 0.454 | 0.357 | 0.626 |
| FCI | Asia | $10^2$ | Clean | -390.1 | 0.188 | 6.5 | 0.273 | 0.188 | 0.5 |
| FCI | Asia | $10^3$ | Clean | -3364.9 | 0.438 | 4.5 | 0.5 | 0.438 | 0.583 |
| FCI | Asia | $10^4$ | Clean | -35362.9 | 0.563 | 3.5 | 0.643 | 0.563 | 0.75 |
| FCI | Asia | $10^5$ | Clean | -349227.8 | 0.688 | 2.5 | 0.786 | 0.688 | 0.917 |
| FCI | Sports | $10^2$ | Clean | -2049.9 | 0.267 | 11 | 0.4 | 0.267 | 0.8 |
| FCI | Sports | $10^3$ | Clean | -16526.7 | 0.4 | 9 | 0.5 | 0.4 | 0.667 |
| FCI | Sports | $10^4$ | Clean | -3774766.6 | 0.567 | 6.5 | 0.567 | 0.567 | 0.567 |
| FCI | Sports | $10^5$ | Clean | -1554611.9 | 0.8 | 3 | 0.8 | 0.8 | 0.8 |
| FCI | Property | $10^2$ | Clean | -5368.2 | 0.242 | 23.5 | 0.349 | 0.242 | 0.625 |
| FCI | Property | $10^3$ | Clean | -43479.2 | 0.319 | 22 | 0.4 | 0.323 | 0.526 |
| FCI | Property | $10^4$ | Clean | -374079.7 | 0.72 | 10.5 | 0.776 | 0.726 | 0.833 |
| FCI | Property | $10^5$ | Clean | -3662583.7 | 0.79 | 6.5 | 0.86 | 0.79 | 0.942 |
| FCI | Alarm | $10^2$ | Clean | -2716.2 | 0.346 | 31 | 0.478 | 0.348 | 0.762 |
| FCI | Alarm | $10^3$ | Clean | -19760.1 | 0.705 | 14.5 | 0.793 | 0.707 | 0.903 |
| FCI | Alarm | $10^4$ | Clean | -157681.5 | 0.837 | 7.5 | 0.885 | 0.837 | 0.939 |
| FCI | Alarm | $10^5$ | Clean | -1553571.0 | 0.89 | 6 | 0.901 | 0.891 | 0.911 |
| FCI | ForMed | $10^2$ | Clean | -7578.6 | 0.195 | 113 | 0.298 | 0.196 | 0.628 |
| FCI | ForMed | $10^3$ | Clean | -68338.0 | 0.333 | 92 | 0.436 | 0.333 | 0.63 |
| FCI | ForMed | $10^4$ | Clean | -648054.0 | 0.496 | 70.5 | 0.593 | 0.496 | 0.737 |
| FCI | ForMed | $10^5$ | Clean | -6444721.0 | 0.524 | 85 | 0.573 | 0.529 | 0.624 |
| FCI | Pathfinder | $10^2$ | Clean | -9534.2 | 0.046 | 198.5 | 0.085 | 0.049 | 0.328 |
| FCI | Pathfinder | $10^3$ | Clean | | | | | | |
| FCI | Pathfinder | $10^4$ | Clean | | | | | | |
| FCI | Pathfinder | $10^5$ | Clean | | | | | | |
| FCI | Asia | $10^2$ | Noisy | -663.6 | 0.017 | 6.5 | 0.125 | 0.083 | 0.25 |
| FCI | Asia | $10^3$ | Noisy | -5930.3 | 0.267 | 5 | 0.4 | 0.333 | 0.5 |
| FCI | Asia | $10^4$ | Noisy | -59353.0 | 0.317 | 6.5 | 0.467 | 0.583 | 0.389 |
| FCI | Asia | $10^5$ | Noisy | -576629.1 | 0.483 | 5.5 | 0.6 | 0.75 | 0.5 |
| FCI | Sports | $10^2$ | Noisy | -2250.1 | 0.269 | 9.5 | 0.412 | 0.269 | 0.875 |
| FCI | Sports | $10^3$ | Noisy | -18268.8 | 0.423 | 7.5 | 0.55 | 0.423 | 0.786 |
| FCI | Sports | $10^4$ | Noisy | -310852.9 | 0.272 | 10 | 0.5 | 0.538 | 0.467 |
| FCI | Sports | $10^5$ | Noisy | -2217174.4 | 0.215 | 11 | 0.5 | 0.615 | 0.421 |
| FCI | Property | $10^2$ | Noisy | -6624.1 | 0.108 | 33 | 0.182 | 0.125 | 0.333 |
| FCI | Property | $10^3$ | Noisy | -68098.8 | 0.427 | 21 | 0.509 | 0.438 | 0.609 |
| FCI | Property | $10^4$ | Noisy | -10648391.5 | 0.616 | 31 | 0.557 | 0.688 | 0.468 |
| FCI | Property | $10^5$ | Noisy | | | | | | |
| FCI | Alarm | $10^2$ | Noisy | -4819.2 | 0.128 | 47.5 | 0.21 | 0.144 | 0.382 |
| FCI | Alarm | $10^3$ | Noisy | -55160.6 | 0.402 | 37 | 0.458 | 0.422 | 0.5 |
| FCI | Alarm | $10^4$ | Noisy | cyclic | 0.579 | 41 | 0.528 | 0.622 | 0.459 |
| FCI | Alarm | $10^5$ | Noisy | cyclic | 0.57 | 62.5 | 0.454 | 0.656 | 0.347 |
| FCI | ForMed | $10^2$ | Noisy | -11914.7 | 0.113 | 139.5 | 0.18 | 0.118 | 0.384 |
| FCI | ForMed | $10^3$ | Noisy | -197110.7 | 0.232 | 120 | 0.327 | 0.236 | 0.532 |
| FCI | ForMed | $10^4$ | Noisy | > 8 parents | 0.364 | 124.5 | 0.407 | 0.375 | 0.445 |
| FCI | ForMed | $10^5$ | Noisy | | | | | | |
| FCI | Pathfinder | $10^2$ | Noisy | -22281.6 | 0.014 | 274.5 | 0.038 | 0.024 | 0.093 |
| FCI | Pathfinder | $10^3$ | Noisy | | | | | | |
| FCI | Pathfinder | $10^4$ | Noisy | | | | | | |
| FCI | Pathfinder | $10^5$ | Noisy | | | | | | |
| GOBNILP | Asia | $10^2$ | Clean | -380.5 | 0.263 | 6.5 | 0.357 | 0.313 | 0.417 |
| GOBNILP | Asia | $10^3$ | Clean | -3252.1 | 0.875 | 1 | 0.933 | 0.875 | 1 |
| GOBNILP | Asia | $10^4$ | Clean | -32468.5 | 1 | 0 | 1 | 1 | 1 |
| GOBNILP | Asia | $10^5$ | Clean | -322699.2 | 1 | 0 | 1 | 1 | 1 |
| GOBNILP | Sports | $10^2$ | Clean | -1952.8 | 0.067 | 14 | 0.118 | 0.067 | 0.5 |

| Algorithm | Case study | Sample size | Data | $BIC_{log2}$ | CPDAG scores BSF | SHD | F1 | Recall | Precision |
|---|---|---|---|---|---|---|---|---|---|
| GOBNILP | Sports | $10^3$ | Clean | -16526.7 | 0.4 | 9 | 0.5 | 0.4 | 0.667 |
| GOBNILP | Sports | $10^4$ | Clean | -157653.0 | 0.319 | 10.5 | 0.44 | 0.367 | 0.55 |
| GOBNILP | Sports | $10^5$ | Clean | -1549920.5 | 0.933 | 1 | 0.933 | 0.933 | 0.933 |
| GOBNILP | Property | $10^2$ | Clean | -5172.6 | 0.216 | 27 | 0.298 | 0.226 | 0.438 |
| GOBNILP | Property | $10^3$ | Clean | -39882.8 | 0.626 | 12.5 | 0.709 | 0.629 | 0.813 |
| GOBNILP | Property | $10^4$ | Clean | -347093.4 | 0.713 | 12.5 | 0.75 | 0.726 | 0.776 |
| GOBNILP | Property | $10^5$ | Clean | -3374328.9 | 0.919 | 2.5 | 0.934 | 0.919 | 0.95 |
| GOBNILP | Alarm | $10^2$ | Clean | -2227.4 | 0.246 | 43 | 0.312 | 0.261 | 0.387 |
| GOBNILP | Alarm | $10^3$ | Clean | -16968.1 | 0.781 | 11 | 0.828 | 0.783 | 0.878 |
| GOBNILP | Alarm | $10^4$ | Clean | -153559.1 | 0.922 | 4.5 | 0.944 | 0.924 | 0.966 |
| GOBNILP | Alarm | $10^5$ | Clean | -1507685.6 | 0.978 | 1 | 0.989 | 0.978 | 1 |
| GOBNILP | ForMed | $10^2$ | Clean | -6969.2 | 0.275 | 153 | 0.324 | 0.29 | 0.367 |
| GOBNILP | ForMed | $10^3$ | Clean | -63159.3 | 0.615 | 69.5 | 0.681 | 0.62 | 0.757 |
| GOBNILP | ForMed | $10^4$ | Clean | -607561.4 | 0.698 | 71.5 | 0.684 | 0.707 | 0.663 |
| GOBNILP | ForMed | $10^5$ | Clean | -7435754.0 | 0.276 | 160.5 | 0.318 | 0.293 | 0.346 |
| GOBNILP | Pathfinder | $10^2$ | Clean | -6338.4 | 0.088 | 230 | 0.137 | 0.097 | 0.229 |
| GOBNILP | Pathfinder | $10^3$ | Clean | -49143.7 | 0.189 | 245 | 0.241 | 0.205 | 0.292 |
| GOBNILP | Pathfinder | $10^4$ | Clean | -633739.9 | 0.121 | 255.5 | 0.168 | 0.136 | 0.221 |
| GOBNILP | Pathfinder | $10^5$ | Clean | | | | | | |
| GOBNILP | Asia | $10^2$ | Noisy | -663.6 | 0.167 | 5 | 0.25 | 0.167 | 0.5 |
| GOBNILP | Asia | $10^3$ | Noisy | -5921.8 | 0.333 | 4 | 0.444 | 0.333 | 0.667 |
| GOBNILP | Asia | $10^4$ | Noisy | -58385.0 | 0.45 | 4.5 | 0.538 | 0.583 | 0.5 |
| GOBNILP | Asia | $10^5$ | Noisy | -575873.9 | 0.467 | 5 | 0.571 | 0.667 | 0.5 |
| GOBNILP | Sports | $10^2$ | Noisy | -1924.8 | 0 | 13 | 0 | 0 | 0 |
| GOBNILP | Sports | $10^3$ | Noisy | -17820.8 | 0.346 | 8.5 | 0.474 | 0.346 | 0.75 |
| GOBNILP | Sports | $10^4$ | Noisy | -167896.2 | 0.423 | 7.5 | 0.524 | 0.423 | 0.688 |
| GOBNILP | Sports | $10^5$ | Noisy | -1655868.2 | 0.885 | 1.5 | 0.885 | 0.885 | 0.885 |
| GOBNILP | Property | $10^2$ | Noisy | -6133.0 | 0.125 | 28 | 0.205 | 0.125 | 0.571 |
| GOBNILP | Property | $10^3$ | Noisy | -52515.7 | 0.325 | 22.5 | 0.412 | 0.328 | 0.553 |
| GOBNILP | Property | $10^4$ | Noisy | -484047.6 | 0.665 | 12.5 | 0.729 | 0.672 | 0.796 |
| GOBNILP | Property | $10^5$ | Noisy | -4683661.8 | 0.654 | 28 | 0.575 | 0.719 | 0.479 |
| GOBNILP | Alarm | $10^2$ | Noisy | -4214.5 | 0.059 | 46 | 0.113 | 0.067 | 0.375 |
| GOBNILP | Alarm | $10^3$ | Noisy | -36141.5 | 0.293 | 35.5 | 0.386 | 0.3 | 0.54 |
| GOBNILP | Alarm | $10^4$ | Noisy | -337495.9 | 0.685 | 21.5 | 0.7 | 0.7 | 0.7 |
| GOBNILP | Alarm | $10^5$ | Noisy | -3288811.5 | 0.621 | 45.5 | 0.535 | 0.678 | 0.442 |
| GOBNILP | ForMed | $10^2$ | Noisy | -10650.3 | 0.089 | 129.5 | 0.151 | 0.089 | 0.481 |
| GOBNILP | ForMed | $10^3$ | Noisy | -98050.7 | 0.204 | 121 | 0.294 | 0.207 | 0.509 |
| GOBNILP | ForMed | $10^4$ | Noisy | -946704.9 | 0.421 | 107 | 0.49 | 0.429 | 0.571 |
| GOBNILP | ForMed | $10^5$ | Noisy | -9350377.2 | 0.507 | 139 | 0.477 | 0.529 | 0.435 |
| GOBNILP | Pathfinder | $10^2$ | Noisy | -12538.5 | 0.044 | 230.5 | 0.08 | 0.046 | 0.339 |
| GOBNILP | Pathfinder | $10^3$ | Noisy | -110941.9 | 0.036 | 259 | 0.07 | 0.043 | 0.182 |
| GOBNILP | Pathfinder | $10^4$ | Noisy | -1015966.8 | 0.143 | 263 | 0.207 | 0.157 | 0.308 |
| GOBNILP | Pathfinder | $10^5$ | Noisy | | | | | | |
| HC | Asia | $10^2$ | Clean | -380.5 | 0.263 | 6.5 | 0.357 | 0.313 | 0.417 |
| HC | Asia | $10^3$ | Clean | -3252.1 | 0.875 | 1 | 0.933 | 0.875 | 1 |
| HC | Asia | $10^4$ | Clean | -32488.5 | 0.763 | 2.5 | 0.765 | 0.813 | 0.722 |
| HC | Asia | $10^5$ | Clean | -322699.2 | 1 | 0 | 1 | 1 | 1 |
| HC | Sports | $10^2$ | Clean | -1952.8 | 0.067 | 14 | 0.118 | 0.067 | 0.5 |
| HC | Sports | $10^3$ | Clean | -16741.4 | 0.219 | 12 | 0.333 | 0.267 | 0.444 |
| HC | Sports | $10^4$ | Clean | -158638.6 | 0.205 | 12.5 | 0.346 | 0.3 | 0.409 |
| HC | Sports | $10^5$ | Clean | -1566465.6 | 0.048 | 16 | 0.323 | 0.333 | 0.313 |
| HC | Property | $10^2$ | Clean | -5172.6 | 0.216 | 27 | 0.298 | 0.226 | 0.438 |
| HC | Property | $10^3$ | Clean | -39950.2 | 0.558 | 15.5 | 0.625 | 0.565 | 0.7 |
| HC | Property | $10^4$ | Clean | -347939.4 | 0.643 | 16.5 | 0.641 | 0.661 | 0.621 |

| Algorithm | Case study | Sample size | Data | $BIC_{log2}$ | CPDAG scores BSF | SHD | F1 | Recall | Precision |
|---|---|---|---|---|---|---|---|---|---|
| HC | Property | $10^5$ | Clean | -3375231.1 | 0.736 | 14.5 | 0.701 | 0.758 | 0.653 |
| HC | Alarm | $10^2$ | Clean | -2256.6 | 0.248 | 42 | 0.316 | 0.261 | 0.4 |
| HC | Alarm | $10^3$ | Clean | -17667.7 | 0.558 | 30.5 | 0.576 | 0.576 | 0.576 |
| HC | Alarm | $10^4$ | Clean | -155683.6 | 0.707 | 25.5 | 0.67 | 0.728 | 0.62 |
| HC | Alarm | $10^5$ | Clean | -1517576.6 | 0.741 | 29.5 | 0.657 | 0.772 | 0.573 |
| HC | ForMed | $10^2$ | Clean | -6995.7 | 0.202 | 150.5 | 0.254 | 0.214 | 0.314 |
| HC | ForMed | $10^3$ | Clean | -63431.8 | 0.416 | 108.5 | 0.461 | 0.424 | 0.504 |
| HC | ForMed | $10^4$ | Clean | -608137.7 | 0.488 | 125.5 | 0.46 | 0.504 | 0.424 |
| HC | ForMed | $10^5$ | Clean | -6012998.5 | 0.62 | 126.5 | 0.527 | 0.641 | 0.447 |
| HC | Pathfinder | $10^2$ | Clean | -6374.7 | 0.08 | 221 | 0.125 | 0.087 | 0.224 |
| HC | Pathfinder | $10^3$ | Clean | -50672.1 | 0.144 | 248 | 0.193 | 0.159 | 0.244 |
| HC | Pathfinder | $10^4$ | Clean | -413791.4 | 0.245 | 240 | 0.287 | 0.262 | 0.319 |
| HC | Pathfinder | $10^5$ | Clean | -3630403.6 | 0.479 | 160.5 | 0.529 | 0.49 | 0.575 |
| HC | Asia | $10^2$ | Noisy | -663.5 | 0.167 | 5 | 0.25 | 0.167 | 0.5 |
| HC | Asia | $10^3$ | Noisy | -5921.8 | 0.333 | 4 | 0.444 | 0.333 | 0.667 |
| HC | Asia | $10^4$ | Noisy | -58410.2 | 0.217 | 6.5 | 0.385 | 0.417 | 0.357 |
| HC | Asia | $10^5$ | Noisy | -576283.2 | 0.167 | 8 | 0.4 | 0.5 | 0.333 |
| HC | Sports | $10^2$ | Noisy | -1924.8 | 0 | 13 | 0 | 0 | 0 |
| HC | Sports | $10^3$ | Noisy | -17844.8 | 0.269 | 9.5 | 0.368 | 0.269 | 0.583 |
| HC | Sports | $10^4$ | Noisy | -168110.7 | 0.423 | 7.5 | 0.524 | 0.423 | 0.688 |
| HC | Sports | $10^5$ | Noisy | -1661218.6 | -0.026 | 14 | 0.308 | 0.308 | 0.308 |
| HC | Property | $10^2$ | Noisy | -6133.0 | 0.125 | 28 | 0.205 | 0.125 | 0.571 |
| HC | Property | $10^3$ | Noisy | -52629.9 | 0.309 | 23 | 0.4 | 0.313 | 0.556 |
| HC | Property | $10^4$ | Noisy | -484310.7 | 0.665 | 12.5 | 0.741 | 0.672 | 0.827 |
| HC | Property | $10^5$ | Noisy | -4700330.7 | 0.754 | 15 | 0.735 | 0.781 | 0.694 |
| HC | Alarm | $10^2$ | Noisy | -4221.9 | 0.059 | 46 | 0.113 | 0.067 | 0.375 |
| HC | Alarm | $10^3$ | Noisy | -36170.5 | 0.258 | 38 | 0.343 | 0.267 | 0.48 |
| HC | Alarm | $10^4$ | Noisy | -338553.7 | 0.508 | 35 | 0.516 | 0.533 | 0.5 |
| HC | Alarm | $10^5$ | Noisy | -3298106.5 | 0.497 | 46.5 | 0.454 | 0.544 | 0.389 |
| HC | ForMed | $10^2$ | Noisy | -10656.8 | 0.082 | 130.5 | 0.14 | 0.082 | 0.479 |
| HC | ForMed | $10^3$ | Noisy | -98117.4 | 0.187 | 122.5 | 0.273 | 0.189 | 0.491 |
| HC | ForMed | $10^4$ | Noisy | -947936.3 | 0.376 | 119 | 0.432 | 0.386 | 0.491 |
| HC | ForMed | $10^5$ | Noisy | -9358595.5 | 0.474 | 146.5 | 0.446 | 0.496 | 0.404 |
| HC | Pathfinder | $10^2$ | Noisy | -12544.6 | 0.044 | 228.5 | 0.081 | 0.046 | 0.35 |
| HC | Pathfinder | $10^3$ | Noisy | -111169.6 | 0.032 | 259 | 0.064 | 0.039 | 0.17 |
| HC | Pathfinder | $10^4$ | Noisy | -1022693.4 | 0.117 | 267 | 0.175 | 0.13 | 0.268 |
| HC | Pathfinder | $10^5$ | Noisy | -9642584.1 | 0.356 | 194 | 0.452 | 0.365 | 0.592 |
| MMHC | Asia | $10^2$ | Clean | -414.9 | 0.188 | 6.5 | 0.273 | 0.188 | 0.5 |
| MMHC | Asia | $10^3$ | Clean | -3535.1 | 0.563 | 3.5 | 0.692 | 0.563 | 0.9 |
| MMHC | Asia | $10^4$ | Clean | -35157.8 | 0.688 | 2.5 | 0.786 | 0.688 | 0.917 |
| MMHC | Asia | $10^5$ | Clean | -349227.8 | 0.688 | 2.5 | 0.786 | 0.688 | 0.917 |
| MMHC | Sports | $10^2$ | Clean | -1952.8 | 0.067 | 14 | 0.118 | 0.067 | 0.5 |
| MMHC | Sports | $10^3$ | Clean | -16692.5 | 0.367 | 9.5 | 0.478 | 0.367 | 0.688 |
| MMHC | Sports | $10^4$ | Clean | -157797.5 | 0.4 | 9 | 0.5 | 0.4 | 0.667 |
| MMHC | Sports | $10^5$ | Clean | -1570734.9 | 0.4 | 9 | 0.5 | 0.4 | 0.667 |
| MMHC | Property | $10^2$ | Clean | -5538.1 | 0.097 | 28 | 0.162 | 0.097 | 0.5 |
| MMHC | Property | $10^3$ | Clean | -44954.2 | 0.323 | 21 | 0.444 | 0.323 | 0.714 |
| MMHC | Property | $10^4$ | Clean | -412077.2 | 0.4 | 19.5 | 0.51 | 0.403 | 0.694 |
| MMHC | Property | $10^5$ | Clean | -4005888.1 | 0.516 | 15 | 0.627 | 0.516 | 0.8 |
| MMHC | Alarm | $10^2$ | Clean | -2549.6 | 0.151 | 40 | 0.237 | 0.152 | 0.538 |
| MMHC | Alarm | $10^3$ | Clean | -19787.4 | 0.435 | 26 | 0.548 | 0.435 | 0.741 |
| MMHC | Alarm | $10^4$ | Clean | -178972.7 | 0.576 | 19.5 | 0.688 | 0.576 | 0.855 |
| MMHC | Alarm | $10^5$ | Clean | -1788917.0 | 0.62 | 17.5 | 0.731 | 0.62 | 0.891 |
| MMHC | ForMed | $10^2$ | Clean | -7559.7 | 0.133 | 122.5 | 0.21 | 0.134 | 0.487 |

| Algorithm | Case study | Sample size | Data | $BIC_{log2}$ | CPDAG scores BSF | SHD | F1 | Recall | Precision |
|---|---|---|---|---|---|---|---|---|---|
| MMHC | ForMed | $10^3$ | Clean | -67410.3 | 0.359 | 88.5 | 0.485 | 0.359 | 0.75 |
| MMHC | ForMed | $10^4$ | Clean | -647318.8 | 0.532 | 66.5 | 0.645 | 0.533 | 0.817 |
| MMHC | ForMed | $10^5$ | Clean | -6577236.5 | 0.524 | 69.5 | 0.644 | 0.525 | 0.833 |
| MMHC | Pathfinder | $10^2$ | Clean | -8546.3 | 0.032 | 196.5 | 0.06 | 0.033 | 0.325 |
| MMHC | Pathfinder | $10^3$ | Clean | -78805.0 | 0.04 | 192 | 0.074 | 0.041 | 0.4 |
| MMHC | Pathfinder | $10^4$ | Clean | -810663.8 | 0.043 | 189.5 | 0.08 | 0.044 | 0.472 |
| MMHC | Pathfinder | $10^5$ | Clean | -7713792.1 | 0.066 | 183 | 0.12 | 0.067 | 0.591 |
| MMHC | Asia | $10^2$ | Noisy | -663.5 | 0.167 | 5 | 0.25 | 0.167 | 0.5 |
| MMHC | Asia | $10^3$ | Noisy | -5921.8 | 0.333 | 4 | 0.444 | 0.333 | 0.667 |
| MMHC | Asia | $10^4$ | Noisy | -58385.0 | 0.45 | 4.5 | 0.538 | 0.583 | 0.5 |
| MMHC | Asia | $10^5$ | Noisy | -576024.1 | 0.55 | 4.5 | 0.643 | 0.75 | 0.563 |
| MMHC | Sports | $10^2$ | Noisy | -1924.8 | 0 | 13 | 0 | 0 | 0 |
| MMHC | Sports | $10^3$ | Noisy | -17820.8 | 0.346 | 8.5 | 0.474 | 0.346 | 0.75 |
| MMHC | Sports | $10^4$ | Noisy | -167896.2 | 0.423 | 7.5 | 0.524 | 0.423 | 0.688 |
| MMHC | Sports | $10^5$ | Noisy | -1716011.0 | 0.654 | 4.5 | 0.739 | 0.654 | 0.85 |
| MMHC | Property | $10^2$ | Noisy | -6153.7 | 0.063 | 30 | 0.114 | 0.063 | 0.667 |
| MMHC | Property | $10^3$ | Noisy | -54179.1 | 0.234 | 24.5 | 0.341 | 0.234 | 0.625 |
| MMHC | Property | $10^4$ | Noisy | -500125.0 | 0.399 | 21 | 0.51 | 0.406 | 0.684 |
| MMHC | Property | $10^5$ | Noisy | -4888455.5 | 0.53 | 19.5 | 0.603 | 0.547 | 0.673 |
| MMHC | Alarm | $10^2$ | Noisy | -4260.0 | 0.043 | 44 | 0.082 | 0.044 | 0.5 |
| MMHC | Alarm | $10^3$ | Noisy | -37114.8 | 0.209 | 36.5 | 0.311 | 0.211 | 0.594 |
| MMHC | Alarm | $10^4$ | Noisy | -344123.4 | 0.478 | 29 | 0.564 | 0.489 | 0.667 |
| MMHC | Alarm | $10^5$ | Noisy | -3436699.7 | 0.445 | 36 | 0.506 | 0.467 | 0.553 |
| MMHC | ForMed | $10^2$ | Noisy | -10753.2 | 0.071 | 130 | 0.126 | 0.071 | 0.526 |
| MMHC | ForMed | $10^3$ | Noisy | -98744.6 | 0.167 | 119.5 | 0.26 | 0.168 | 0.573 |
| MMHC | ForMed | $10^4$ | Noisy | -956135.7 | 0.346 | 105 | 0.447 | 0.35 | 0.62 |
| MMHC | ForMed | $10^5$ | Noisy | -9514725.3 | 0.427 | 108 | 0.506 | 0.436 | 0.604 |
| MMHC | Pathfinder | $10^2$ | Noisy | -13276.5 | 0.008 | 230 | 0.017 | 0.009 | 0.333 |
| MMHC | Pathfinder | $10^3$ | Noisy | -116795.2 | 0.021 | 240.5 | 0.043 | 0.024 | 0.212 |
| MMHC | Pathfinder | $10^4$ | Noisy | -1227246.0 | 0.039 | 223 | 0.073 | 0.039 | 0.563 |
| MMHC | Pathfinder | $10^5$ | Noisy | -12022006.9 | 0.051 | 223 | 0.094 | 0.052 | 0.5 |
| PC-Stable | Asia | $10^2$ | Clean | -390.1 | 0.188 | 6.5 | 0.273 | 0.188 | 0.5 |
| PC-Stable | Asia | $10^3$ | Clean | -3364.9 | 0.438 | 4.5 | 0.5 | 0.438 | 0.583 |
| PC-Stable | Asia | $10^4$ | Clean | -35157.8 | 0.688 | 2.5 | 0.786 | 0.688 | 0.917 |
| PC-Stable | Asia | $10^5$ | Clean | -349227.8 | 0.688 | 2.5 | 0.786 | 0.688 | 0.917 |
| PC-Stable | Sports | $10^2$ | Clean | -2049.9 | 0.267 | 11 | 0.4 | 0.267 | 0.8 |
| PC-Stable | Sports | $10^3$ | Clean | -20105.5 | 0.5 | 7.5 | 0.6 | 0.5 | 0.75 |
| PC-Stable | Sports | $10^4$ | Clean | -3774766.6 | 0.567 | 6.5 | 0.567 | 0.567 | 0.567 |
| PC-Stable | Sports | $10^5$ | Clean | -1555121.5 | 0.6 | 6 | 0.6 | 0.6 | 0.6 |
| PC-Stable | Property | $10^2$ | Clean | -5368.2 | 0.258 | 23 | 0.364 | 0.258 | 0.615 |
| PC-Stable | Property | $10^3$ | Clean | -43553.6 | 0.365 | 21.5 | 0.451 | 0.371 | 0.575 |
| PC-Stable | Property | $10^4$ | Clean | -374056.4 | 0.671 | 12 | 0.737 | 0.677 | 0.808 |
| PC-Stable | Property | $10^5$ | Clean | -3662583.7 | 0.79 | 6.5 | 0.86 | 0.79 | 0.942 |
| PC-Stable | Alarm | $10^2$ | Clean | -2599.1 | 0.337 | 30.5 | 0.477 | 0.337 | 0.816 |
| PC-Stable | Alarm | $10^3$ | Clean | -18341.1 | 0.685 | 14.5 | 0.768 | 0.685 | 0.875 |
| PC-Stable | Alarm | $10^4$ | Clean | -169771.3 | 0.748 | 12.5 | 0.784 | 0.75 | 0.821 |
| PC-Stable | Alarm | $10^5$ | Clean | -1535554.0 | 0.922 | 4.5 | 0.934 | 0.924 | 0.944 |
| PC-Stable | ForMed | $10^2$ | Clean | -7631.7 | 0.202 | 112 | 0.313 | 0.203 | 0.683 |
| PC-Stable | ForMed | $10^3$ | Clean | -67991.0 | 0.413 | 81 | 0.54 | 0.413 | 0.781 |
| PC-Stable | ForMed | $10^4$ | Clean | -647193.9 | 0.525 | 66.5 | 0.628 | 0.525 | 0.78 |
| PC-Stable | ForMed | $10^5$ | Clean | -6477511.0 | 0.585 | 75.5 | 0.642 | 0.591 | 0.703 |
| PC-Stable | Pathfinder | $10^2$ | Clean | | | | | | |
| PC-Stable | Pathfinder | $10^3$ | Clean | | | | | | |
| PC-Stable | Pathfinder | $10^4$ | Clean | | | | | | |

| Algorithm | Case study | Sample size | Data | $BIC_{log2}$ | CPDAG scores BSF | SHD | F1 | Recall | Precision |
|---|---|---|---|---|---|---|---|---|---|
| PC-Stable | Pathfinder | $10^5$ | Clean | | | | | | |
| PC-Stable | Asia | $10^2$ | Noisy | -663.6 | 0.017 | 6.5 | 0.125 | 0.083 | 0.25 |
| PC-Stable | Asia | $10^3$ | Noisy | -5930.3 | 0.267 | 5 | 0.4 | 0.333 | 0.5 |
| PC-Stable | Asia | $10^4$ | Noisy | -59345.5 | 0.317 | 6.5 | 0.467 | 0.583 | 0.389 |
| PC-Stable | Asia | $10^5$ | Noisy | -576629.1 | 0.483 | 5.5 | 0.6 | 0.75 | 0.5 |
| PC-Stable | Sports | $10^2$ | Noisy | -2250.1 | 0.269 | 9.5 | 0.412 | 0.269 | 0.875 |
| PC-Stable | Sports | $10^3$ | Noisy | -17833.7 | 0.385 | 8 | 0.5 | 0.385 | 0.714 |
| PC-Stable | Sports | $10^4$ | Noisy | -200722.7 | 0.349 | 9 | 0.571 | 0.615 | 0.533 |
| PC-Stable | Sports | $10^5$ | Noisy | -2022272.6 | 0.264 | 10.5 | 0.576 | 0.731 | 0.475 |
| PC-Stable | Property | $10^2$ | Noisy | -6595.9 | 0.13 | 30.5 | 0.209 | 0.141 | 0.409 |
| PC-Stable | Property | $10^3$ | Noisy | -80059.0 | 0.455 | 21 | 0.536 | 0.469 | 0.625 |
| PC-Stable | Property | $10^4$ | Noisy | -20657247.4 | 0.635 | 29.5 | 0.577 | 0.703 | 0.489 |
| PC-Stable | Property | $10^5$ | Noisy | | | | | | |
| PC-Stable | Alarm | $10^2$ | Noisy | -4712.4 | 0.119 | 47 | 0.197 | 0.133 | 0.375 |
| PC-Stable | Alarm | $10^3$ | Noisy | -83978.7 | 0.513 | 32 | 0.565 | 0.533 | 0.6 |
| PC-Stable | Alarm | $10^4$ | Noisy | -1597266.4 | 0.634 | 38.5 | 0.575 | 0.678 | 0.5 |
| PC-Stable | Alarm | $10^5$ | Noisy | > 8 parents | 0.67 | 58 | 0.523 | 0.756 | 0.4 |
| PC-Stable | ForMed | $10^2$ | Noisy | -11931.4 | 0.123 | 140 | 0.195 | 0.129 | 0.4 |
| PC-Stable | ForMed | $10^3$ | Noisy | > 8 parents | 0.246 | 119 | 0.348 | 0.25 | 0.574 |
| PC-Stable | ForMed | $10^4$ | Noisy | > 8 parents | 0.467 | 111 | 0.517 | 0.479 | 0.563 |
| PC-Stable | ForMed | $10^5$ | Noisy | | | | | | |
| PC-Stable | Pathfinder | $10^2$ | Noisy | -19287.0 | 0.019 | 271.5 | 0.045 | 0.028 | 0.116 |
| PC-Stable | Pathfinder | $10^3$ | Noisy | | | | | | |
| PC-Stable | Pathfinder | $10^4$ | Noisy | | | | | | |
| PC-Stable | Pathfinder | $10^5$ | Noisy | | | | | | |
| SaiyanH | Asia | $10^2$ | Clean | -385.8 | 0.337 | 6.5 | 0.467 | 0.438 | 0.5 |
| SaiyanH | Asia | $10^3$ | Clean | -3256.6 | 0.65 | 4 | 0.75 | 0.75 | 0.75 |
| SaiyanH | Asia | $10^4$ | Clean | -32468.5 | 1 | 0 | 1 | 1 | 1 |
| SaiyanH | Asia | $10^5$ | Clean | -322699.2 | 1 | 0 | 1 | 1 | 1 |
| SaiyanH | Sports | $10^2$ | Clean | -2037.1 | -0.057 | 17 | 0.174 | 0.133 | 0.25 |
| SaiyanH | Sports | $10^3$ | Clean | -16526.7 | 0.4 | 9 | 0.5 | 0.4 | 0.667 |
| SaiyanH | Sports | $10^4$ | Clean | -157653.0 | 0.319 | 10.5 | 0.44 | 0.367 | 0.55 |
| SaiyanH | Sports | $10^5$ | Clean | -1549920.5 | 0.833 | 2.5 | 0.833 | 0.833 | 0.833 |
| SaiyanH | Property | $10^2$ | Clean | -5364.4 | 0.278 | 30.5 | 0.333 | 0.306 | 0.365 |
| SaiyanH | Property | $10^3$ | Clean | -40993.6 | 0.597 | 17 | 0.655 | 0.613 | 0.704 |
| SaiyanH | Property | $10^4$ | Clean | -360836.3 | 0.739 | 9 | 0.78 | 0.742 | 0.821 |
| SaiyanH | Property | $10^5$ | Clean | -3510295.0 | 0.723 | 9.5 | 0.75 | 0.726 | 0.776 |
| SaiyanH | Alarm | $10^2$ | Clean | -2292.9 | 0.294 | 44.5 | 0.354 | 0.315 | 0.403 |
| SaiyanH | Alarm | $10^3$ | Clean | -17250.8 | 0.689 | 18 | 0.736 | 0.696 | 0.78 |
| SaiyanH | Alarm | $10^4$ | Clean | -155209.0 | 0.812 | 10.5 | 0.824 | 0.815 | 0.833 |
| SaiyanH | Alarm | $10^5$ | Clean | -1509362.8 | 0.946 | 2.5 | 0.967 | 0.946 | 0.989 |
| SaiyanH | ForMed | $10^2$ | Clean | -7081.1 | 0.2 | 148 | 0.253 | 0.21 | 0.319 |
| SaiyanH | ForMed | $10^3$ | Clean | -63997.7 | 0.475 | 85 | 0.559 | 0.478 | 0.673 |
| SaiyanH | ForMed | $10^4$ | Clean | -614154.9 | 0.629 | 57 | 0.699 | 0.63 | 0.784 |
| SaiyanH | ForMed | $10^5$ | Clean | -6080414.0 | 0.68 | 48 | 0.755 | 0.681 | 0.847 |
| SaiyanH | Pathfinder | $10^2$ | Clean | -32031.4 | 0.125 | 274 | 0.165 | 0.144 | 0.193 |
| SaiyanH | Pathfinder | $10^3$ | Clean | -83292.0 | 0.193 | 281 | 0.223 | 0.215 | 0.232 |
| SaiyanH | Pathfinder | $10^4$ | Clean | -459824.5 | 0.199 | 218 | 0.263 | 0.21 | 0.35 |
| SaiyanH | Pathfinder | $10^5$ | Clean | -4043389.2 | 0.301 | 215.5 | 0.349 | 0.315 | 0.392 |
| SaiyanH | Asia | $10^2$ | Noisy | -699.9 | -0.1 | 9 | 0.167 | 0.167 | 0.167 |
| SaiyanH | Asia | $10^3$ | Noisy | -5950.9 | 0.367 | 5 | 0.5 | 0.5 | 0.5 |
| SaiyanH | Asia | $10^4$ | Noisy | -58458.5 | 0.383 | 5.5 | 0.538 | 0.583 | 0.5 |
| SaiyanH | Asia | $10^5$ | Noisy | -577562.1 | 0.533 | 4 | 0.615 | 0.667 | 0.571 |
| SaiyanH | Sports | $10^2$ | Noisy | -2134.1 | -0.008 | 13.5 | 0.25 | 0.192 | 0.357 |

| Algorithm | Case study | Sample size | Data | $BIC_{log2}$ | BSF | SHD | F1 | Recall | Precision |
|---|---|---|---|---|---|---|---|---|---|
| | | | | | CPDAG scores | | | | |
| SaiyanH | Sports | $10^3$ | Noisy | -17888.6 | 0.279 | 9.5 | 0.45 | 0.346 | 0.643 |
| SaiyanH | Sports | $10^4$ | Noisy | -167896.2 | 0.423 | 7.5 | 0.524 | 0.423 | 0.688 |
| SaiyanH | Sports | $10^5$ | Noisy | -1660280.8 | 0.262 | 10 | 0.5 | 0.462 | 0.545 |
| SaiyanH | Property | $10^2$ | Noisy | -6575.9 | 0.083 | 44.5 | 0.158 | 0.141 | 0.18 |
| SaiyanH | Property | $10^3$ | Noisy | -53198.5 | 0.304 | 28.5 | 0.368 | 0.328 | 0.42 |
| SaiyanH | Property | $10^4$ | Noisy | -485262.8 | 0.643 | 15 | 0.7 | 0.656 | 0.75 |
| SaiyanH | Property | $10^5$ | Noisy | -4723117.3 | 0.78 | 11.5 | 0.785 | 0.797 | 0.773 |
| SaiyanH | Alarm | $10^2$ | Noisy | -4516.8 | 0.14 | 58 | 0.203 | 0.178 | 0.235 |
| SaiyanH | Alarm | $10^3$ | Noisy | -36508.4 | 0.397 | 34.5 | 0.468 | 0.411 | 0.544 |
| SaiyanH | Alarm | $10^4$ | Noisy | -339272.9 | 0.586 | 31.5 | 0.604 | 0.611 | 0.598 |
| SaiyanH | Alarm | $10^5$ | Noisy | -3304616.6 | 0.564 | 43.5 | 0.514 | 0.611 | 0.444 |
| SaiyanH | ForMed | $10^2$ | Noisy | -11040.9 | 0.11 | 172.5 | 0.157 | 0.125 | 0.211 |
| SaiyanH | ForMed | $10^3$ | Noisy | -98493.8 | 0.249 | 132 | 0.321 | 0.257 | 0.429 |
| SaiyanH | ForMed | $10^4$ | Noisy | -948967.7 | 0.37 | 115 | 0.436 | 0.379 | 0.515 |
| SaiyanH | ForMed | $10^5$ | Noisy | -9385561.6 | 0.433 | 111 | 0.49 | 0.443 | 0.549 |
| SaiyanH | Pathfinder | $10^2$ | Noisy | -14746.5 | 0.015 | 312.5 | 0.045 | 0.033 | 0.073 |
| SaiyanH | Pathfinder | $10^3$ | Noisy | -112929.6 | 0.038 | 299.5 | 0.075 | 0.054 | 0.121 |
| SaiyanH | Pathfinder | $10^4$ | Noisy | -1038257.5 | 0.22 | 250 | 0.296 | 0.235 | 0.4 |
| SaiyanH | Pathfinder | $10^5$ | Noisy | -9712041.2 | 0.303 | 232 | 0.374 | 0.317 | 0.456 |
| TABU | Asia | $10^2$ | Clean | -380.5 | 0.263 | 6.5 | 0.357 | 0.313 | 0.417 |
| TABU | Asia | $10^3$ | Clean | -3252.1 | 0.875 | 1 | 0.933 | 0.875 | 1 |
| TABU | Asia | $10^4$ | Clean | -32468.5 | 1 | 0 | 1 | 1 | 1 |
| TABU | Asia | $10^5$ | Clean | -322699.2 | 1 | 0 | 1 | 1 | 1 |
| TABU | Sports | $10^2$ | Clean | -1952.8 | 0.067 | 14 | 0.118 | 0.067 | 0.5 |
| TABU | Sports | $10^3$ | Clean | -16667.2 | 0.319 | 10.5 | 0.458 | 0.367 | 0.611 |
| TABU | Sports | $10^4$ | Clean | -157653.0 | 0.319 | 10.5 | 0.44 | 0.367 | 0.55 |
| TABU | Sports | $10^5$ | Clean | -1566465.6 | 0.048 | 16 | 0.323 | 0.333 | 0.313 |
| TABU | Property | $10^2$ | Clean | -5172.6 | 0.216 | 27 | 0.298 | 0.226 | 0.438 |
| TABU | Property | $10^3$ | Clean | -39947.8 | 0.571 | 16 | 0.632 | 0.581 | 0.692 |
| TABU | Property | $10^4$ | Clean | -347939.4 | 0.643 | 16.5 | 0.641 | 0.661 | 0.621 |
| TABU | Property | $10^5$ | Clean | -3374482.7 | 0.884 | 4.5 | 0.887 | 0.887 | 0.887 |
| TABU | Alarm | $10^2$ | Clean | -2256.6 | 0.248 | 42 | 0.316 | 0.261 | 0.4 |
| TABU | Alarm | $10^3$ | Clean | -17186.4 | 0.675 | 20.5 | 0.708 | 0.685 | 0.733 |
| TABU | Alarm | $10^4$ | Clean | -154473.7 | 0.851 | 11.5 | 0.849 | 0.859 | 0.84 |
| TABU | Alarm | $10^5$ | Clean | -1507808.7 | 0.897 | 7.5 | 0.874 | 0.902 | 0.847 |
| TABU | ForMed | $10^2$ | Clean | -6993.5 | 0.212 | 153 | 0.261 | 0.225 | 0.31 |
| TABU | ForMed | $10^3$ | Clean | -63400.3 | 0.416 | 109.5 | 0.455 | 0.424 | 0.492 |
| TABU | ForMed | $10^4$ | Clean | -607622.4 | 0.5 | 122 | 0.472 | 0.514 | 0.436 |
| TABU | ForMed | $10^5$ | Clean | -6010410.2 | 0.63 | 116.5 | 0.544 | 0.649 | 0.469 |
| TABU | Pathfinder | $10^2$ | Clean | -6374.7 | 0.08 | 221 | 0.125 | 0.087 | 0.224 |
| TABU | Pathfinder | $10^3$ | Clean | -50407.0 | 0.143 | 255 | 0.188 | 0.159 | 0.231 |
| TABU | Pathfinder | $10^4$ | Clean | -410954.9 | 0.257 | 241.5 | 0.296 | 0.274 | 0.32 |
| TABU | Pathfinder | $10^5$ | Clean | -3583196.4 | 0.504 | 160.5 | 0.545 | 0.515 | 0.578 |
| TABU | Asia | $10^2$ | Noisy | -663.5 | 0.167 | 5 | 0.25 | 0.167 | 0.5 |
| TABU | Asia | $10^3$ | Noisy | -5921.8 | 0.333 | 4 | 0.444 | 0.333 | 0.667 |
| TABU | Asia | $10^4$ | Noisy | -58385.0 | 0.533 | 4 | 0.615 | 0.667 | 0.571 |
| TABU | Asia | $10^5$ | Noisy | -575902.1 | 0.633 | 4 | 0.667 | 0.833 | 0.556 |
| TABU | Sports | $10^2$ | Noisy | -1924.8 | 0 | 13 | 0 | 0 | 0 |
| TABU | Sports | $10^3$ | Noisy | -17833.7 | 0.385 | 8 | 0.5 | 0.385 | 0.714 |
| TABU | Sports | $10^4$ | Noisy | -168110.7 | 0.423 | 7.5 | 0.524 | 0.423 | 0.688 |
| TABU | Sports | $10^5$ | Noisy | -1661218.6 | -0.026 | 14 | 0.308 | 0.308 | 0.308 |
| TABU | Property | $10^2$ | Noisy | -6133.0 | 0.125 | 28 | 0.205 | 0.125 | 0.571 |
| TABU | Property | $10^3$ | Noisy | -52533.7 | 0.321 | 23.5 | 0.404 | 0.328 | 0.525 |
| TABU | Property | $10^4$ | Noisy | -484310.7 | 0.665 | 12.5 | 0.741 | 0.672 | 0.827 |

| Algorithm | Case study | Sample size | Data | $\mathrm{BIC}_{\log 2}$ | CPDAG scores BSF | SHD | F1 | Recall | Precision |
|---|---|---|---|---|---|---|---|---|---|
| TABU | Property | $10^5$ | Noisy | -4697235.7 | 0.759 | 17.5 | 0.718 | 0.797 | 0.654 |
| TABU | Alarm | $10^2$ | Noisy | -4221.9 | 0.059 | 46 | 0.113 | 0.067 | 0.375 |
| TABU | Alarm | $10^3$ | Noisy | -36170.5 | 0.258 | 38 | 0.343 | 0.267 | 0.48 |
| TABU | Alarm | $10^4$ | Noisy | -337848.8 | 0.573 | 33 | 0.563 | 0.6 | 0.529 |
| TABU | Alarm | $10^5$ | Noisy | -3290052.1 | 0.494 | 54 | 0.427 | 0.556 | 0.347 |
| TABU | ForMed | $10^2$ | Noisy | -10650.3 | 0.089 | 129.5 | 0.151 | 0.089 | 0.481 |
| TABU | ForMed | $10^3$ | Noisy | -98099.8 | 0.19 | 122 | 0.277 | 0.193 | 0.491 |
| TABU | ForMed | $10^4$ | Noisy | -947348.1 | 0.361 | 123 | 0.413 | 0.371 | 0.464 |
| TABU | ForMed | $10^5$ | Noisy | -9351785.0 | 0.471 | 144 | 0.444 | 0.493 | 0.404 |
| TABU | Pathfinder | $10^2$ | Noisy | -12544.6 | 0.044 | 228.5 | 0.081 | 0.046 | 0.35 |
| TABU | Pathfinder | $10^3$ | Noisy | -111169.6 | 0.032 | 259 | 0.064 | 0.039 | 0.17 |
| TABU | Pathfinder | $10^4$ | Noisy | -1022368.9 | 0.141 | 275 | 0.202 | 0.157 | 0.283 |
| TABU | Pathfinder | $10^5$ | Noisy | -9623277.0 | 0.362 | 203 | 0.447 | 0.374 | 0.555 |

# References

Bouckaert, R. (1994). Properties of Bayesian belief network learning algorithms. In *Proceedings of the 10th Conference on Uncertainty in Artificial Intelligence* (UAI 1994), pp. 102–109.

Bouchaert, R. (1995). Bayesian belief networks: from construction to inference. Ph.D thesis, University of Utrecht.

Center for Causal Discovery. (2020). Tetrad manual. University of Pittsburgh, USA. [Online] http://cmu-phil.github.io/tetrad/manual/

Chen, Y. and Tian, J. (2014) Finding the *k*-best Equivalence Classes of Bayesian Network structures for Model Averaging. In *Proceedings of the 28th AAAI Conference on Artificial Intelligence*, Vol. 28, Iss. 1.

Chickering, D. (2002). Learning equivalence classes of Bayesian-network structures. *Journal of Machine Learning Research*, Vol. 2, pp. 445–498.

Colombo, D., and Maathuis, M. H. (2014). Order-Independent Constraint-Based Causal Structure Learning. *Journal of Machine Learning Research*, Vol. 15, pp 3921–3962.

Constantinou, A. (2019). The Bayesys user manual. Queen Mary University of London, London, UK. [Online]. Available: http://bayesian-ai.eecs.qmul.ac.uk/bayesys/

Constantinou, A. (2019b). Evaluating structure learning algorithms with a balanced scoring function. *arXiv:1905.12666* [cs.LG].

Constantinou, A. C. (2020). Learning Bayesian networks that enable full propagation of evidence. *IEEE Access*, Vol. 8, pp. 124845–123856.

Constantinou, A. C., Liu, Y., Chobtham, K., Guo, Z., and Kitson, N. K. (2020). The Bayesys data and Bayesian network repository. Queen Mary University of London, London, UK. [Online]. Available: http://bayesian-ai.eecs.qmul.ac.uk/bayesys/

Constantinou, A. C., Liu, Y., Chobtham, K., Guo, Z., and Kitson, N. K. (2021). Large-scale empirical validation of Bayesian Network structure learning algorithms with noisy data. *International Journal of Approximate Reasoning*, Vol. 131, pp. 151–188.

Cussens, J., (2011). Bayesian network learning with cutting planes. In *Proceedings of the 27th Conference on Uncertainty in Artificial Intelligence* (UAI 2011), pp. 153–160.

Cussens, J., and Bartlett, M. (2015). GOBNILP 1.6.2 User/Developer Manual. University of York, UK. [Online]. Available: https://www.cs.york.ac.uk/aig/sw/gobnilp/manual.pdf

de Campos, C.P., Zeng, Z. and Ji, Q. (2009). Structure learning of Bayesian networks using constraints. In *Proceedings of the 26th Annual International Conference on Machine Learning*, pp. 113–120.

Goudie, R., and Mukherjee, S. (2016). A Gibbs sampler for learning DAGs. *Journal of Machine Learning Research*, Vol. 17, pp. 1–39.

Guo, Z. and Constantinou, A.C. (2020). Approximate learning of high dimensional Bayesian network structures via pruning of Candidate Parent Sets. *Entropy*, Vol. 22, Iss. 10, Article 1142.

Heckerman, D., Geiger, D., and Chickering, D. (1995). Learning Bayesian networks: the combination of knowledge and statistical data. *Machine Learning*, vol. 20, pp. 197–243.

Kitson, K. K., Constantinou, A. C., Guo, Z., Liu, Y., and Chobtham, K. (2021). A survey of Bayesian network structure learning. *arXiv:2109.11415* [cs.LG]

Kuipers, J., Suter, P., and Moffa, G. (2021). Efficient Sampling and Structure Learning of Bayesian Networks. *arXiv:1803.07859* [stat.ML]

Madigan, D., Andersson, S.A., Perlman, M.D. and Volinsky, C.T. (1996). Bayesian model averaging and model selection for Markov equivalence classes of acyclic digraphs. *Communications in Statistics - Theory and Methods*, Vol. 25, Iss. 11, pp. 2493–2519.

Rantanen, K., Hyttinen, A., and Järvisalo, M. (2021). Maximal Ancestral Graph Structure Learning via Exact Search. In *Proceedings of the 37th Conference on Uncertainty in Artificial Intelligence (UAI 2021)*, In Press.

Richardson, T. and Spirtes, P. (2002). Ancestral graph Markov models. *Annals of Statistics*, Vol. 30, Iss. 4, pp. 962–1030.

Scanagatta, M., de Campos, C.P., Corani, G. and Zaffalon, M. (2015). Learning Bayesian networks with thousands of variables. In *Advances in Neural Information Processing Systems (NIPS)*, pp. 1864–1872.

Scutari, M. (2021). Package 'bnlearn'. University of Oxford, England, UK. [Online] https://www.bnlearn.com/documentation/bnlearn-manual.pdf

Spirtes, P., and Glymour, C. (1991). An algorithm for fast recovery of sparse causal graphs. *Social Science Computer Review*, Vol. 9, Iss. 1.

Spirtes, P., Meek, C., and Richardson, T. (1999). An algorithm for causal inference in the presence of latent variables and selection bias. In *Clark Glymour and Gregory Cooper (Eds.), Computation, Causation, and Discovery*. The MIT Press, Cambridge, MA, pp. 211–252.

Spirtes, P., Glymour, C., and Scheines, R. (2000). Causation, Prediction, and Search (2nd ed.). The MIT Press. Cambridge, Massachusetts and London, England.

Tsamardinos, I., Brown, L. E., and Aliferis, C. F. (2006). The Max-Min Hill-Climbing Bayesian Network Structure Learning Algorithm. *Machine Learning*, Vol. 65, pp. 31–78.

Tsirlis, K., Lagani, V., Triantafillou, S., and Tsamardinos, I. (2018). On scoring maximal ancestral graphs with the max-min hill climbing algorithm. *International Journal of Approximate Reasoning*, Vol. 102, pp. 74–85.